# Comparative study of motion detection methods for video surveillance systems

**Kamal Sehairi,[a] Chouireb Fatima,[a] Jean Meunier,[b]**
[a] University of Laghouat Amar Telidji, TSS Laboratory, Laghouat, Algeria 03000
[b] University of Montreal, Department of Computer Science and Operations Research, Montreal, Canada

**Abstract**. The objective of this study is to compare several change detection methods for a mono static camera and identify the best method for different complex environments and backgrounds in indoor and outdoor scenes. To this end, we used the CDnet video dataset* as a benchmark that consists of many challenging problems, ranging from basic simple scenes to complex scenes affected by bad weather and dynamic backgrounds. Twelve change detection methods, ranging from simple temporal differencing to more sophisticated methods, were tested and several performance metrics were used to precisely evaluate the results. Because most of the considered methods have not previously been evaluated on this recent large scale dataset, this work compares these methods to fill a lack in the literature, and thus this evaluation joins as complementary compared with the previous comparative evaluations. Our experimental results show that there is no perfect method for all challenging cases; each method performs well in certain cases and fails in others. However, this study enables the user to identify the most suitable method for his or her needs.

**Address all correspondence to:** Kamal Sehairi, University of Laghouat Amar Telidji, TSS Laboratory, Laghouat, Algeria 03000; Tel: +213 661-931-582; E-mail: k.sehairi@lagh-univ.dz; sehairikamel@yahoo.fr

## 1 Introduction

Motion detection, which is the fundamental step in video surveillance, aims to detect regions corresponding to moving objects. The resultant information often forms the basis for higher-level operations that require well-segmented results, such as object classification and action or activity recognition. However, motion detection suffers from problems caused by source noise, complex backgrounds, variations in scene illumination, and the shadows of static and moving objects. Various methods have been proposed to overcome these problems by retaining only the moving object of interest. These methods are classified[1-3] into three major categories: background subtraction,[4,5] temporal differencing,[6,7] and optical flow[8,9]. Temporal differencing is highly adaptive to dynamic environments; however, it generally exhibits poor performance in extracting all relevant feature pixels. Therefore, techniques such as morphological operations and hole filling are applied to effectively extract the shape of a moving object. Background subtraction provides the most complete feature data, but is extremely sensitive to dynamic scene changes due to lighting and extraneous events. Several background modelling methods have been proposed to overcome these problems. Bouwmans[10] classified these advanced methods into seven categories: basic background modelling[11,12,13], statistical background modelling[10,14-16], fuzzy background modelling[17,18], background clustering[19,20], neural network background modelling[21-24], wavelet background modelling[25,26], and background estimation[27,28,29]. Furthermore, Goyette *et al.*[30] categorized background modelling techniques into six families: basic[31-34], parametric[14,15,19,35-37], non-parametric and data-driven[16,38-41], matrix decomposition[42-45], motion segmentation[46-48] and machine learning[21,22,49-52]. Sobral and Vacavant[12] adopted a similar categorization for motion detection methods: basic (frame difference, mean and variance over time)[18,54-57], statistical[14,15,58-60], fuzzy[17,18,61-63], neural and neuro-fuzzy[21,22,64,65] and other models (PCA[42],Vumeter[66]). Optical flow can be used to detect independently moving objects even in the presence of camera motion. However, most optical flow methods are computationally complex and cannot be applied to full-frame video streams in real time without specialized hardware[13]. Recent categories have emerged in the last few years such as: advanced non parametric modelling (Vibe[39], PBAS[40], Vibe+[68]), background modelling by decomposition into matrices[69-74], background modelling by decomposition into tensors[75-77].

* www.ChangeDetection.net

# 2 Related works

## *2.1 Survey papers*

Several surveys on motion detection methods have been presented in the last decade, the authors tried to detail the algorithms used, categorize them and explain their different steps such as post-processing, initialization, background modelling and foreground generation. Bouwmans[78] presented the most complete survey of traditional and recent methods for foreground detection with more than 300 references, these methods were categorized by the mathematical approach used. In addition, the author explored the different datasets available, existing background subtraction libraries and codes. Elhabian *et al.*[79] provided a detailed explanation of different background modelling methods. Specifically, they explained how models can be updated and initialized, and explored ways to measure and evaluate their performance. Morris *et al.*[80] reviewed three pixel-wise subtraction techniques and compared their capabilities in seven points, resource utilization, computational speed, robustness to noise, precision of the output (no numerical results were provided), complexity of the environment, level of autonomy and scalability. Bouwmans[81] in another work presented a review paper in which he classifies the different improvement techniques made to principal component analysis method (Eigen Backgrounds) and compared these techniques with Gaussian detection methods and kernel density estimation using Wallflower datasets. Cuevas *et al.*[82] presented a state of the art of different techniques for detecting stationary foreground objects e.g abandoned luggage, people remaining temporarily static or objects removed from the background, the authors addressed the main challenges in this field with different datasets available for testing these methods. Bux *et.al*[83] gave a complete survey on human activity recognition (HAR), providing the state of the art of foreground segmentation, feature extraction and activity recognition, which constitute the different phases of HAR systems, in foreground segmentation phase, the authors first categorized all methods to background construction-based segmentation for static cameras, and foreground extraction-based segmentation for moving cameras, they detailed the different steps of background construction (Initialization, maintenance and foreground detection) and classified these methods into five models, basic, statistical, fuzzy, neural network and others. Cristani *et al.*[84] presented a comprehensive review of background subtraction techniques for mono and multi-sensor surveillance systems, that considers different kind of sensors (visible, infrared and audio). The authors presented a different taxonomy, classifying motion detection methods into three categories: per-pixel, per-region, per-frame processing, and from each category emerges sub-categories. The authors propose solutions for different challenging situations (adopted from the Wallflower dataset[85]) using the fusion of multi-sensors.

*2.2 comparative papers*

In recent years, many studies have also attempted to compare different motion detection methods. The aim of these studies is to define the accuracy, the speed, memory requirements and capabilities to handle several situations. For this purpose, different challenging datasets have been developed in order to give a fair benchmark for all methods. Table 1 summarizes some previous comparison studies, the evaluated methods, datasets and performance metrics used for each comparison.

**Table 1 Comparison studies on motion detection methods**

| Comparative studies | Tested methods | Datasets used | Metrics used |
|---|---|---|---|
| Toyama *et al.*[86] | Frame difference[86]<br>mean + threshold[86]<br>Mean + covariance (Running Gaussian Average)[14]<br>Mixture of Gaussians[15]<br>Normalized block correlation[87]<br>Temporal derivative (MinMax)[88] | Wallflower dataset[85] | False negative (FN)<br>False positive (FP) |
| | Bayesian decision[89]<br>Subspace learning-principle component analysis(Eigen-Backgrounds)[42]<br>Linear predictive filter[86]<br>Wallflower method[86] | | |
| Picardi[4] | Running Gaussian average[14]<br>Temporal median filter[90,91]<br>Mixture of Gaussians[15]<br>Kernel density estimation (KDE)[16]<br>Sequential kernel density approximation[92]<br>Subspace learning-principle component analysis(Eigen-Backgrounds)[85]<br>Co-occurrence of image variations[93] | / | Limited accuracy (*L*)<br>Intermediate accuracy (*M*)<br>High accuracy (*H*) |
| Cheung *et al.*[94] | Frame differencing[86,94]<br>Temporal median filter[90,91]<br>Linear predictive filter[86]<br>Kernel density estimation(KDE)[16]<br>Approximated median filter[94]<br>Kalman filter[95]<br>Mixture of Gaussians[15] | KOGS/-IAKS Universitaet Karlsruhe dataset[96] | Recall<br>Precision |
| Benezeth *et al.*[97] | Temporal median filter (basic motion detection)[90,91]<br>Running Gaussian average (one Gaussian)[14]<br>Minimum, maximum and maximum Inter-frame difference (MinMax)[88]<br>Mixture of Gaussians[15]<br>Kernel density estimation (KDE)[16]<br>Codebook[19]<br>Subspace learning-principle component analysis (Eigen-Backgrounds)[89] | Synthetic videos<br>Semi-synthetic videos and VSSN 2006 dataset[98]<br>IBM dataset[99] | Recall<br>Precision |
| Bouwmans[10] | Mixture of Gaussians (MoG)[15]<br>Mixture of Gaussians with particle swarm optimization (MoG-PSO)[100]<br>Improved MoG[101]<br>MoG with MRF[102]<br>MoG improved HLS color space[103]<br>Spatial-Time adaptive per pixel mixture of Gaussian (S-TAP-MoG)[104]<br>Adaptive spatio-temporal neighborhood analysis (ASTNA)[105]<br>Subspace learning-principle component analysis (Eigen-Backgrounds)[42]<br>Subspace learning independent component analysis (SL-ICA)[106]<br>Subspace learning incremental non-negative matrix factorization (SL-INMF)[107]<br>Subspace learning using incremental rank-tensor (SL-IRT)[108] | Wallflower dataset[85] | False negative (FN)<br>False positive (FP) |
| Goyette *et al.*[109] | Euclidean distance[97]<br>Mahalanobis distance[97]<br>Local-self similarity[110]<br>Mixture of Gaussians (MoG)[15]<br>GMM KaewTraKulPong[58]<br>GMM Zivkovic[60]<br>GMM RECTGAUSS-Tex[111]<br>Bayesian multi-layer[36] | CDnet 2012 dataset[115] | Recall<br>Specificity<br>False Positive Rate<br>False Negative Rate<br>Percentage of wrong classifications<br>Precision<br>F-measure |
| | ViBe[39]<br>Kernel density estimation(KDE)[16]<br>KDE Nonaka *et al.*[112]<br>KDE Yoshinaga *et al.*[113]<br>Self-organized Background Subtraction (SOBS)[21]<br>Spatially coherent self-organized background subtraction (SC-SOBS)[22]<br>Chebyshev probability[28]<br>ViBe+[66]<br>Probabilistic super-pixel Markov random fields (PSP-MRF)[114]<br>Pixel-based adaptive segmenter (PBAS)[40] | | |
| Wang *et al.*[116] | Euclidean distance[97]<br>Mahalanobis distance[97]<br>Multiscale spatio-temp BG Model[117]<br>GMM Zivkovic[60]<br>CP3-online[118]<br>Mixture of Gaussians (MoG)[15]<br>Kernel density estimation(KDE)[16]<br>Spatially coherent self-organized background subtraction (SC-SOBS)[22]<br>K-nearest neighbor method (KNN)[60,119]<br>Fast self-tuning BS[120]<br>Spectral-360[121]<br>Weightless neural networks (CwisarDH)[51,122]<br>Majority Vote-all[116]<br>Self-balanced local sensitivity (SuBSENSE)[123]<br>Flux tensor with split Gaussian models (FTSG)[124]<br>Majority Vote-3[116] | CDnet 2014 dataset[76] | Recall<br>Specificity<br>False Positive Rate<br>False Negative Rate<br>Percentage of wrong classifications<br>Precision<br>F-measure |
| Jodoin *et al.*[126] | Euclidean distance[97]<br>Mahalanobis distance[97]<br>Mixture of Gaussians (MoG)[15]<br>GMM Zivkovic[60]<br>GMM KaewTraKulPong[58]<br>GMM RECTGAUSS-Tex[111]<br>Kernel density estimation (KDE)[16]<br>KDE Nonaka *et al.*[112]<br>KDE Yoshinaga *et al.*[113]<br>Self-organized background subtraction (SOBS)[21]<br>Spatially coherent self-organized background subtraction (SC-SOBS)[22]<br>K-nearest neighbor method (KNN)[119]<br>Spectral-360[121]<br>Flux tensor with split Gaussian models (FTSG)[124]<br>Pixel-based adaptive segmenter (PBAS)[40]<br>Probabilistic super-pixel Markov random fields (PSP-MRF)[114]<br>Splitting Gaussian mixture model (SGMM)[127]<br>Splitting over-dominating modes GMM (SGMM-SOD)[128]<br>Dirichlet process GMM (DPGMM)[129] | CDnet 2012 dataset[115] | False Positive Rate<br>False Negative Rate<br>Percentage of wrong classifications |
| | Bayesian multi-layer[36]<br>Histogram over time[13]<br>Local-self similarity[110] | | |
| Bianco *et al.*[130] | IUTIS-1[130]<br>IUTIS-2[130]<br>IUTIS-3[130]<br>Flux tensor with split Gaussian models (FTSG)[124]<br>Self-balanced local sensitivity (SuBSENSE)[123]<br>Weightless neural networks (CwisarDH)[51,122]<br>Spectral-360[121]<br>fast self-tuning BS[120]<br>K-nearest neighbor method (KNN)[119]<br>Kernel density estimation(KDE)[16]<br>Spatially coherent self-organized background subtraction (SC-SOBS)[22]<br>Euclidean distance[97]<br>Mahalanobis distance[97]<br>Multiscale spatio-temp BG model[117]<br>CP3-online[118]<br>Mixture of Gaussians (MoG)[15]<br>GMM Zivkovic[60]<br>Fuzzy spatial coherence-based SOBS[65]<br>Region-based mixture of Gaussians (RMoG)[131] | CDnet 2014 dataset[125] | Recall<br>Specificity<br>False Positive Rate<br>False Negative Rate<br>Percentage of wrong classifications<br>Precision<br>F-measure |
| Xu *et al*[132] | Mixture of Gaussians (MoG)[15]<br>Kernel density estimation(KDE)[16]<br>Codebook[19]<br>Self-organized background subtraction (SOBS)[21]<br>ViBe[39]<br>Pixel-based adaptive segmenter (PBAS)[40]<br>GMM Zivkovic[60] (adaptive GMM)<br>sample consensus (SACON)[133,134] | CDnet 2014 dataset[125]<br>Video dataset proposed in Wen *et al.*[135] | Recall<br>Specificity<br>False Positive Rate<br>False Negative Rate<br>Percentage of wrong classifications<br>Precision<br>F-measure |

The table above shows more than 60 motion detection methods that were tested on 8 datasets. Other datasets exist like: CMU[†136], UCSD[‡137], BMC[§138], SCOVIS[139], MarDCT[**140], moreover, new datasets were introduced with depth camera such as: Kinect database[††141] and RGB-D object detection dataset[‡‡142]. We can also find more specialized video datasets such as

† http://www.cs.cmu.edu/~yaser/#software
‡ http://www.svcl.ucsd.edu/projects/background_subtraction/ucsdbgsub_dataset.htm
§ http://bmc.iut-auvergne.com/
** http://www.dis.uniroma1.it/~labrococo/MAR/index.htm
†† https://imatge.upc.edu/web/resources/kinect-database-foreground-segmentation
‡‡ http://eis.bristol.ac.uk/~mc13306/

Fish4knowledge[§§143] for underwater fish detection and tracking. Other comparative studies can be found in[144-148].

Owing to the importance of the motion detection step, it is necessary to examine other motion detection methods that have not been evaluated thus far. In particular, various simple modifications to original methods (pre-processing, thresholding, filtering, etc.) can lead to different results.

The objective of this study is to evaluate and compare different motion detection methods and identify the best method for different situations using a challenging complete dataset. To this end, we tested the following methods: temporal differencing (frame difference)[86,149,150], three-frame difference (3FD)[151-153], adaptive background (average filter)[90,154,155], forgetting morphological temporal gradient (FMTG)[156], $\Sigma\Delta$ Background estimation[157,158], spatio-temporal Markov field[159-161], running Gaussian average (RGA)[14,162,163], mixture of Gaussians (MoG)[15,59,164], spatio-temporal entropy image (STEI)[165,166], difference-based spatio-temporal entropy image (DSTEI)[166,167], eigen-background (Eig-Bg)[42,168,169] and simplified self-organized map (Simp-SOBS)[24] methods. Many of these methods (3FD, $\Sigma\Delta$, FMTG, STEI, DSTEI, Simp-SOBS) have not been previously evaluated on challenging dataset, to this end, we used the CDnet2012[30,115] and CDnet2014[125] datasets, and compared them with the well-known and classical algorithm of motion detection in the literature (RGA, MoG and Eig-Bg). The CDnet2014[***] comprises a total of 53 videos of indoor and outdoor scenes with more than 159000 images. Each scene represents different moving objects, such as boats, cars, trucks, cyclists and pedestrians, captured in different scenarios (baseline, shadow, intermittent object motion) as well as under challenging conditions (bad weather, camera jitter, dynamic background and thermal). For each video, a ground truth is provided to allow precise and unified comparison of the change detection methods. Furthermore, the following seven metrics were used for evaluation: recall, specificity, false positive rate, false negative rate, percentage of wrong classification, precision and F-measure.

[§§] http://groups.inf.ed.ac.uk/f4k/index.html
[***] http://wordpress-jodoin.dmi.usherb.ca/dataset2014/

The remainder of this paper is organized as follows. Section 3 reviews the motion detection algorithms used in this study (frame difference techniques, background modelling techniques and energy-based methods). Section 4 provides a detailed explanation of the evaluation metrics used to score and evaluate the above-mentioned methods. Section 5 presents and discusses the results for different categories. Finally, Section 6 concludes the paper.

# 3 Motion Detection Methods

Motion detection by a fixed camera poses a major challenge for video surveillance systems in terms of extracting the shape of moving objects. This is due to several problems related to the monitored environment, such as complex backgrounds (e.g. tree leaf movement), weather conditions (e.g. snow or rain) and variations in illumination, as well as the characteristics of the moving object itself, such as the similarity of its color to the background color, its size and its distance from the camera. Therefore, in recent years, several methods have been developed to overcome these problems. This section reviews the motion detection algorithms used in this comparative study.

## *3.1 Frame Difference (Temporal Differencing)*

The frame difference method is the simplest method for detecting temporal changes in intensity in video frames. In a grey-level image, for each pixel with coordinates $(x, y)$ in frame $I_{t-1}$, we compute the absolute difference with its corresponding coordinates in the next frame $I_t$ as follows:

$$\zeta(x,y)=\left|I_t(x,y)-I_{t-1}(x,y)\right| \quad (1)$$

For an RGB color image, we can compute this difference by various means, such as Manhattan distance (2), Euclidean distance (3), or Chebyshev distance (4) [97,170-173]

$$\zeta(x,y)=\left|I_t^R(x,y)-I_{t-1}^R(x,y)\right|+\left|I_t^G(x,y)-I_{t-1}^G(x,y)\right|+\left|I_t^B(x,y)-I_{t-1}^B(x,y)\right| \quad (2)$$

$$\zeta(x,y)=\sqrt{\left(I_t^R(x,y)-I_{t-1}^R(x,y)\right)^2+\left(I_t^G(x,y)-I_{t-1}^G(x,y)\right)^2+\left(I_t^B(x,y)-I_{t-1}^B(x,y)\right)^2} \quad (3)$$

$$\zeta(x,y)=\max\left\{\left|I_t^R(x,y)-I_{t-1}^R(x,y)\right|,\left|I_t^G(x,y)-I_{t-1}^G(x,y)\right|,\left|I_t^B(x,y)-I_{t-1}^B(x,y)\right|\right\} \quad (4)$$

where $I_t^C(x,y)$ represents the pixel value in the $C$ channel.

In spite of its simplicity, this method offers the following advantages. It exhibits good performance in dynamic environments (e.g. during sunrise or under cloud cover) and works well at the standard video frame rate. In addition, the algorithm is easy to implement, with relatively low design complexity, and can be executed effectively when applied to a real-time system[174].

### *3.2 Three-frame Difference*

The three-frame difference[151] method is based on the temporal differencing method. Two frame difference operations given by (5) and (6) are performed; then, the results are thresholded using (7) and combined using (8), i.e. the AND logical operator (or the minimum) (see Fig. 1).

$$\zeta_1(x,y)=\left|I_t(x,y)-I_{t-1}(x,y)\right| \quad (5)$$

$$\zeta_2(x,y)=\left|I_t(x,y)-I_{t+1}(x,y)\right| \quad (6)$$

$$\psi_t(x,y)=\begin{cases}0 & if\ \ \zeta_t(x,y)<Th_t \quad background \\ 1 & otherwise \quad foreground\end{cases} \quad (7)$$

$$\zeta_t(x,y)=Min\left(\psi_1(x,y),\psi_2(x,y)\right) \quad (8)$$

This method is robust to noise and provides good detection results for slow moving objects.

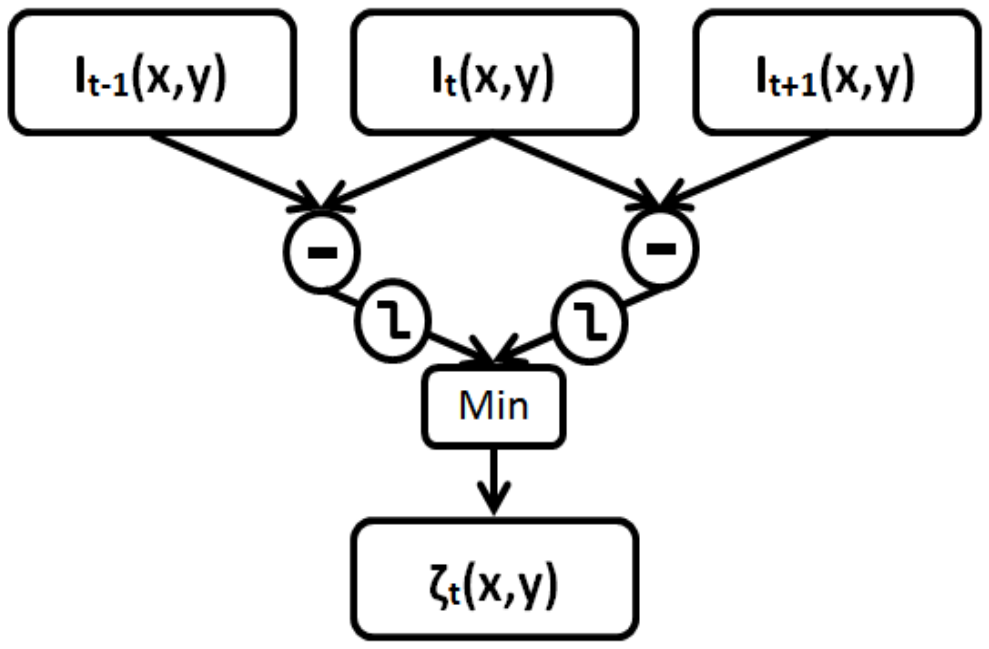

**Fig. 1.** Three-frame difference method

### *3.3 Adaptive Background Subtraction (Running Average Filter)*

The concept underlying this method is to compute the average of the previous $N$ frames to model the background, to update the first background image by considering new static objects in the scene. The background image is obtained as in Ref.[155], where $\tau$ is the time required to acquire $N$ images.

$$B(x,y)=\frac{1}{\tau}\sum_{t=1}^{\tau} I_t(x,y) \tag{9}$$

From (9), this method consumes a significant amount of memory, which causes problems for real-time implementation in particular. To overcome these problems, it is better to compute the background recursively (Fig. 2) as follows:

$$B_{t+1}(x,y)=(1-\alpha)B_t(x,y)+\alpha I_t(x,y) \tag{10}$$

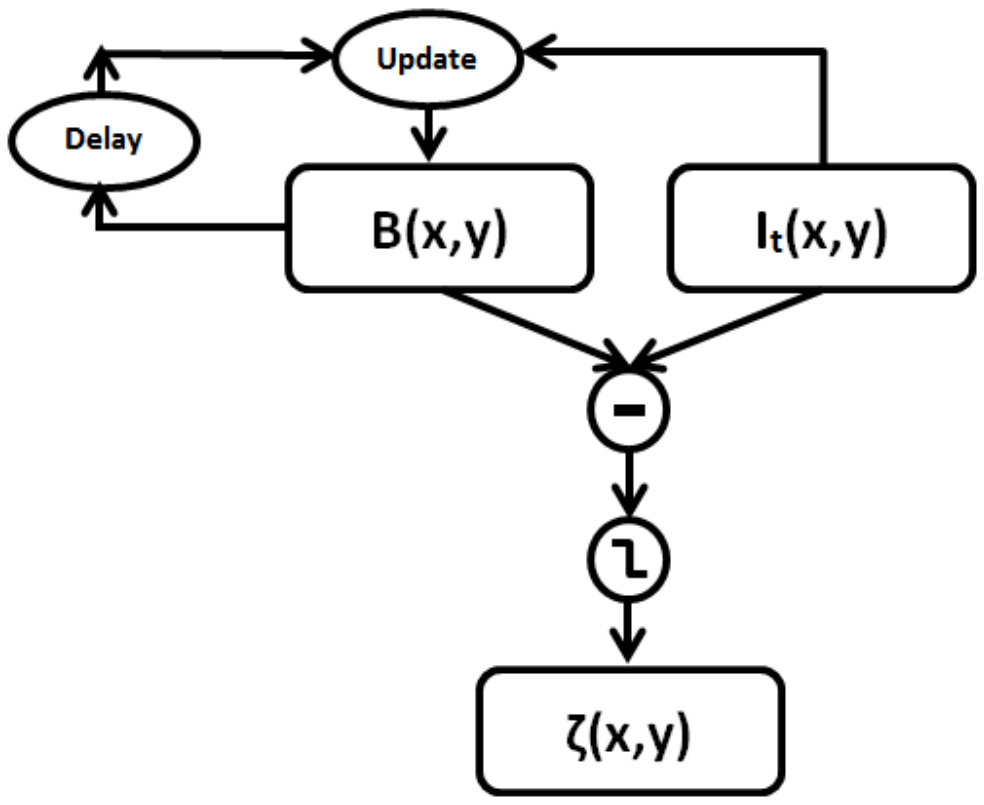

**Fig. 2.** Adaptive background detection

where $\alpha \in [0,1]$ is a time constant that specifies how fast new information supplants old observations. The larger the value of $\alpha$, the higher the rate at which the background frame is updated with new changes in the scene. However, $\alpha$ cannot be too large, because it may cause artificial 'tails' behind moving objects[154]. In fact, to prevent tail formation, $\alpha$ must be fixed according to the observed scene, the size and speed of the moving objects, and the distance of these objects from the camera. Furthermore, the problem of continuous movement of small background objects, especially in outdoor scenes (e.g. fluttering flags and swaying tree branches), can be addressed by segmenting such objects with the moving objects.

### *3.4 Forgetting Morphological Temporal Gradient*

In this method, which was first introduced by Richefeu *et al.*[156], the difference between temporal dilation and temporal erosion defines the change, given by

$$\delta_\tau \left( I_t(x,y) \right) = \max_{z \in \tau} \left\{ I_{t+z}(x,y) \right\} \tag{11}$$

$$\varepsilon_\tau \left( I_t(x,y) \right) = \min_{z \in \tau} \left\{ I_{t+z}(x,y) \right\} \tag{12}$$

where $\tau = [t_1, t_2]$ is the temporal structuring element.

In order to reduce not only the memory consumption linked to the use of the temporal structuring element but also the sensitivity of this method to large sudden variations, the authors used a running average filter to recursively estimate the values of the temporal erosion and dilation. Thus, (11) and (12) respectively become

$$M_t(x,y) = \alpha I_t(x,y) + (1-\alpha) \max \left\{ I_t(x,y), M_{t-1}(x,y) \right\} \tag{13}$$

$$m_t(x,y) = \alpha I_t(x,y) + (1-\alpha) \min \left\{ I_t(x,y), m_{t-1}(x,y) \right\} \tag{14}$$

where $M_t(x,y)$, $m_t(x,y)$ denote the forgetting temporal dilation and the forgetting temporal erosion, respectively.

The forgetting morphological temporal gradient is given by

$$\Gamma_t(x,y) = M_t(x,y) - m_t(x,y) \tag{15}$$

Further, the authors tried to combine this method with the $\Sigma\Delta$ filter to improve the results and automatically define the time constant $\alpha$.

### 3.5 $\Sigma\Delta$ *Background Estimation*

Proposed by Manzanera *et al.*[157], this method is based on the non-linear $\Sigma\Delta$ filter used in electronics applications for analogue-to-digital conversion. The principle of this algorithm is to estimate two values, namely the current background image $M_t$ and the time-variance image $V_t$, using an iterative process to increment or decrement these values. The algorithm is executed in four steps[158]:

(1) Computation of $\Sigma\Delta$ mean

$$\left.\begin{aligned} & \mathrm{M}_0(x,y) = I_0(x,y) \\ & M_t(x,y) = M_{t-1}(x,y) + \operatorname{sgn}\left(I_t(x,y) - M_{t-1}(x,y)\right) \end{aligned}\right\} \tag{16}$$

(2) Computation of difference

$$\Delta_t(x,y) = \left|M_t(x,y) - I_t(x,y)\right| \tag{17}$$

(3) Computation of $\Sigma\Delta$ variance

$$\left.\begin{aligned} & \mathrm{V}_0(x,y) = \Delta_0(x,y) \\ & \text{if } \Delta_t(x,y) \neq 0, \mathrm{V}_t(x,y) = \mathrm{V}_{t-1}(x,y) + \operatorname{sgn}\left(N \times \Delta_t(x,y) - \mathrm{V}_{t-1}(x,y)\right) \end{aligned}\right\} \tag{18}$$

(4) Computation of motion label

$$\begin{cases} \mathrm{D}_t\,(\text{x,y})=0 \;\; \text{if } \Delta_t(x,y) < \mathrm{V}_t(x,y) \\ \mathrm{D}_t\,(\text{x,y})=1 \;\; \text{else} \end{cases} \tag{19}$$

The only parameter to be set in this method is *N*, which represents the amplification factor. However, the application of this method entails several problems such as noise and ghost effects due to moving objects that remain static for long periods in the scene. To overcome this problem, the authors proposed a hybrid geodesic morphological reconstruction filter[157] based on the forgetting morphological operator[156], and given by

$$\Delta_t' = H\ \mathrm{Rec}_{\alpha}^{\Delta_t}\left(Min\left(\left\|\nabla\left(I_t\right)\right\|,\left\|\nabla\left(\Delta_t\right)\right\|\right)\right) \tag{20}$$

where the gradients of $I_t$ and $\Delta_t$ are obtained by convolution with Sobel masks, $\alpha$ is time constant. Further, the classical geodesic relaxation $\mathrm{Rec}^{\Delta_t}$ is defined by the geodesic dilation as $\mathrm{Rec}^{\Delta_t}\left(Min\left(\left\|\nabla\left(I_t\right)\right\|,\left\|\nabla\left(\Delta_t\right)\right\|\right)\right) = Min\left(\delta_B\left(Min\left(\left\|\nabla\left(I_t\right)\right\|,\left\|\nabla\left(\Delta_t\right)\right\|\right),\Delta_t\right)\right)$, where $\delta$ is the morphological dilation operator, and $B$ is the structuring element.

### *3.6 Markov Random Field (MRF)-based Motion Detection Algorithm*

Introduced by Bouthemy and Lalande[159], this algorithm aims to improve image difference using a Markovian process. To this end, the authors have defined motion detection in images as a binary labelling problem, where the appropriate labels are given by

$$\begin{cases} e(x,y,t) = 1 \text{ if the pixel belongs to a moving object} \\ e(x,y,t) = 0 \text{ if the pixel belongs to a static background} \end{cases} \tag{21}$$

and the observation is the absolute difference between two consecutive frames or between the current frame and a reference image:

$$O_t(x,y) = \left|I_t(x,y) - I_{t-1}(x,y)\right| \tag{22}$$

The maximum *a posteriori* (MAP) criterion is used to estimate the appropriate labels of field $E$ given field of observation $O$ interpreted by maximizing the conditional probability

$$\max_e P\left[E = e \middle| O = o\right] \tag{23}$$

Using Bayes' theorem, this is equivalent to

$$\max_e \frac{P\left[O = o \middle| \mathrm{E} = e\right] \mathrm{P}\left[E = e\right]}{P\left[O = o\right]} \tag{24}$$

where $P\left[O = o\right]$ is constant with respect to the maximization because the observations are inputs. Conversely, the maximization of $P\left[O = o \middle| \mathrm{E} = e\right]\mathrm{P}\left[E = e\right]$ is equivalent to the minimization of an energy function derived from the Hammersley-Clifford theorem, which states that Markov random fields exhibit a Gibbs distribution with an energy function as follows[175,176]:

$$P\left[E = e \middle| O = o\right] = \frac{e^{-U(o,e)}}{Z} \tag{25}$$

where $Z$ is a normalizing factor. The energy function U is given by the sum of two terms,

$$U(e,o) = U_m(e) + U_a(o,e) \tag{26}$$

where $U_m$ denotes the energy that ensures spatio-temporal homogeneity and $U_a$ denotes the adequacy energy that ensures good coherence of the solution compared to the observed data:

$$U_a(o,e) = \frac{1}{2\sigma^2}\left[o - \psi(e)\right]^2, \psi(e) = \begin{cases} 0 & \text{if } \left|I_t(x,y) - I_{t\text{-}1}(x,y)\right| < Th \\ \alpha & \text{else} \end{cases} \tag{27}$$

$$U_m(e) = \sum_{c \in C} V_c(e_s, e_r) \tag{28}$$

where $c$ denotes a set of binary cliques associated with the chosen neighborhood system describing spatio-temporal interactions between the different pixel intensities[159]. In our case, there is a 3×3 spatial neighborhood window and two temporal connections: *(x,y,t)* to *(x,y,t-1)* and *(x,y,t)* to *(x,y,t+1)* (see Fig. 3).

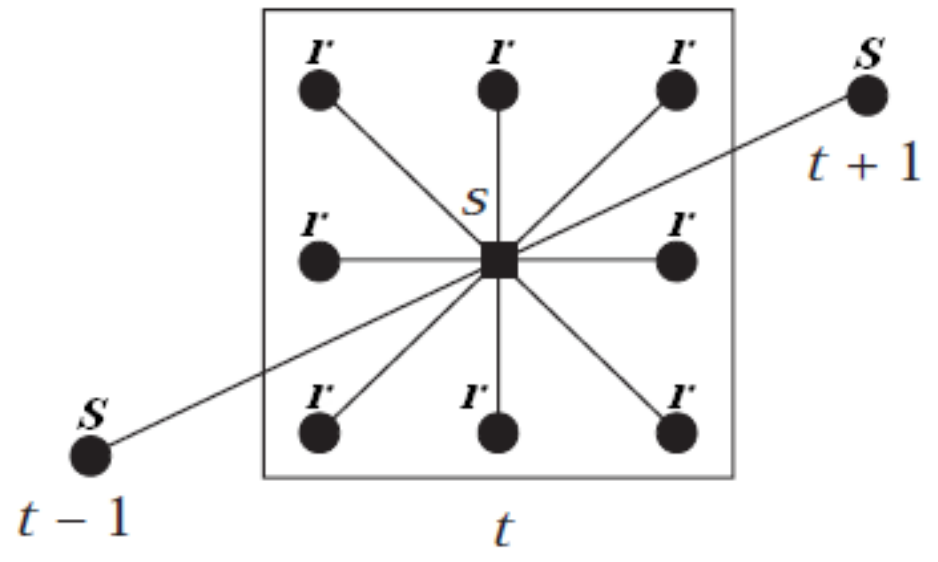

**Fig. 3.** 3×3 Spatio-temporal neighborhood[160].

Further, $V_c$ is given by

$$\begin{cases} V_c(e_s,e_r) = V_s(e_s,e_r) + V_p(\mathrm{e}_s^t,\mathrm{e}_s^{t-1}) + V_p(\mathrm{e}_s^t,\mathrm{e}_s^{t+1}) \\ V_s(e_s,e_r) = \begin{cases} -\beta_s & \text{if } \mathrm{e}_s = e_r \\ +\beta_s & \text{if } \mathrm{e}_s \neq e_r \end{cases} \\ V_p(\mathrm{e}_s^t,\mathrm{e}_s^{t-1}) = \begin{cases} -\beta_p & \text{if } \mathrm{e}_s^t = \mathrm{e}_s^{t-1} \\ +\beta_p & \text{if } \mathrm{e}_s^t \neq \mathrm{e}_s^{t-1} \end{cases} \\ V_p(\mathrm{e}_s^t,\mathrm{e}_s^{t+1}) = \begin{cases} -\beta_f & \text{if } \mathrm{e}_s^t = \mathrm{e}_s^{t+1} \\ +\beta_f & \text{if } \mathrm{e}_s^t \neq \mathrm{e}_s^{t+1} \end{cases} \end{cases} \tag{29}$$

After defining the energy $U$ for our Markovian model, the authors considered the problem of minimizing this energy, ultimately using an iterative deterministic relaxation technique[159] (iterated conditional mode method) (see Fig. 4).

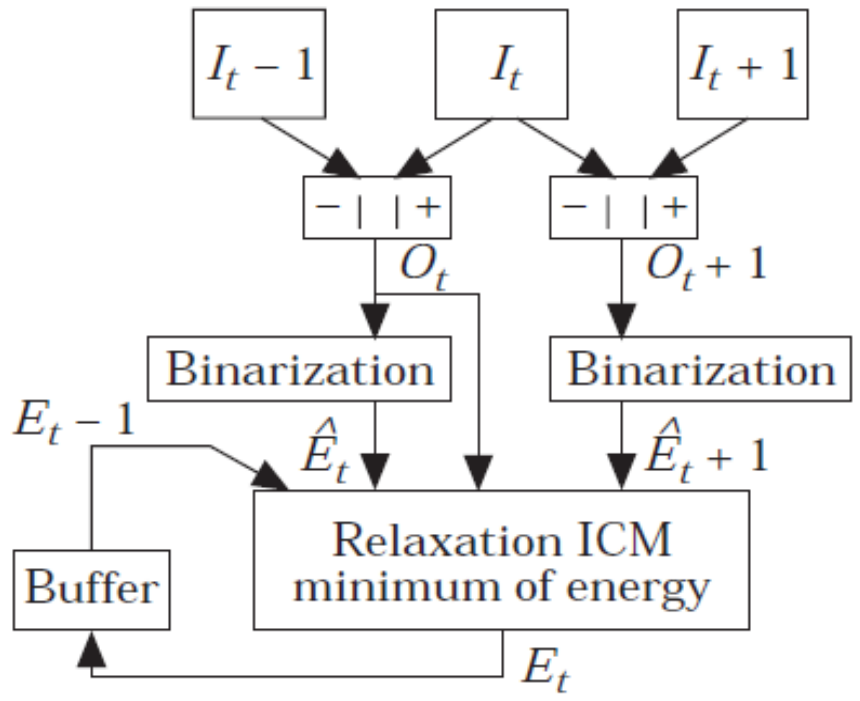

**Fig. 4.** MRF-based motion detection[160]

*3.7 Running Gaussian Average (One Gaussian)*[14]

In this method, the background is modelled by fitting a Gaussian distribution $(\mu,\sigma)$ over a histogram for each pixel[4]; this gives the probability density function (pdf) of the background. In order to update this pdf, a running average filter is applied to the parameters of the Gaussian:

$$\mu_t = \alpha I_t + (1-\alpha)\mu_{t-1} \tag{30}$$

$$\sigma_t^2 = \alpha\left(I_t - \mu_t\right)^2 + (1-\alpha)\sigma_{t-1}^2 \tag{31}$$

Then, the pixels correspond to a moving object if the following inequality is satisfied:

$$\left|I_t - \mu_t\right| > D\sigma_t \tag{32}$$

where $D$ is the deviation threshold (e.g. $D = 2.5$).

This method offers the advantages of high execution speed and low memory consumption. However, it suffers from problems associated with the use of the running average (appropriate choices of $\alpha$ and deviation threshold $D$). Moreover, the use of a single Gaussian to model the background will not give good results for complex backgrounds; in addition, it will favor the extraction of the shadows of moving objects.

*3.8 Gaussian Mixture Model (GMM)*

In the Gaussian mixture model (or mixture of Gaussians MoG), proposed by Stauffer and Grimson[15], the temporal histogram of each pixel $X$ is modelled using a mixture of $K$ Gaussian distributions in order to precisely model a dynamic background. For example, the periodic or random oscillation of a tree branch that sways in the wind and hides the sun is modelled using two Gaussians. One Gaussian models the temporal variation in the intensity of the pixels when the tree branch obstructs the sun, and the other Gaussian represents the different local intensities produced by the sun. The intensity of each pixel is compared to these Gaussian mixtures, which represent the probability distribution of possible intensities belonging to the dynamic background model. Low probability of belonging to these Gaussian mixtures indicates that the pixel belongs to a moving object. The probability of observing the current pixel value in the multidimensional case is given by

$$P\left(X_t\right) = \sum_{k=1}^{K} \omega_{k,t} \eta\left(\mu_{k,t}, \Sigma_{k,t}, X_t\right) \tag{33}$$

where $\omega_{k,t}$ is the estimated weight associated with the $k^{\text{th}}$ Gaussian at time *t*, $\mu_{k,t}$ is the mean of the $k^{\text{th}}$ Gaussian at time *t*, and $\Sigma_{k,t}$ is the covariance matrix. Further, $\eta$ is a Gaussian probability density function:

$$\eta\left(\mu_{k,t},\Sigma_{k,t},X_t\right)=\frac{1}{(2\pi)^{\frac{n}{2}}\left|\Sigma\right|^{\frac{1}{2}}}e^{-\frac{1}{2}(X_t-\mu_t)^T\Sigma^{-1}(X_t-\mu_t)} \tag{34}$$

Owing to limited memory and computing capacity, the authors have set *K* in the range of 3 to 5, and they have assumed that the RGB color components are independent and have the same variances. Hence, the covariance matrix is of the form[59]

$$\Sigma_{i,k}=\sigma_{i,k}^2 I \tag{35}$$

The first step is to initialize the parameters of the Gaussians ($\omega_k,\mu_k,\Sigma_k$). Then, a test is performed to match each new pixel *X* with the existing Gaussians.

$$\left|\mu_k-X_t\right|\le D\sigma_k \qquad \text{k=1,...,M} \tag{36}$$

where *M* is the number of Gaussians and $D$ is the deviation threshold.

If a match is found, we update the parameters of this matched Gaussian as follows:

$$\rho=\frac{\alpha}{\omega} \tag{37}$$

$$\omega_t=(1-\alpha)\omega_{t-1}+\alpha \tag{38}$$

$$\mu_t=\rho X_t+(1-\rho)\mu_{t-1} \tag{39}$$

$$\sigma_t^2=\rho\left(X_t-\mu_t\right)^2+(1-\rho)\sigma_{t-1}^2 \tag{40}$$

where $\alpha$ and $\rho$ are learning rates; here, $\rho$ is taken from the alternative approximation proposed by Power and Schoonees[177]. For the other unmatched distributions, we maintain their mean and variance; we update only the weights as in (41).

$$\omega_t=(1-\alpha)\omega_{t-1} \tag{41}$$

Then, we normalize all the weights as $\omega_k/\sum_{k=1}^{M}\omega_k$ .

If no match is found with any of the *K* distributions, we create a new distribution that replaces the parameters of the least probable one, with the current pixel value as its mean, an initially high variance, and low prior weight:

$$\mu_t = X_t \quad (42)$$

$$\sigma_t^2 = \sigma_0^2 \text{ (largest value)} \quad (43)$$

$$\omega_t = \min(\omega_t) \ \left(\text{smallest value}\right) \quad (44)$$

Then, to distinguish the foreground distribution from the background distribution, we order the distributions by the ratio of their weights to their standard deviations ($\omega_k / \sigma_k$), assuming that the higher and more compact the distribution, the greater the likelihood of belonging to the background. Then, the first *B* distributions in the ranking order satisfying (45) are considered background[4]:

$$\sum_{k=1}^{B} \omega_k > T \quad (45)$$

where *T* is a threshold value. Finally, each new pixel value *X* is compared to these background distributions. If a match is found, this pixel is considered to be a background pixel; otherwise, it is considered to be a foreground pixel.

$$\left|\mu_k - X_t\right| \le D\sigma_k \qquad \text{k=1,...,B} \quad (46)$$

This method can yield good results for dynamic backgrounds by fitting multiple Gaussians to represent the background more effectively. However, it entails two problems: high computational complexity and parameter initialization. Furthermore, additional parameters must be fixed, such as the threshold value and learning values $\alpha$ and $\rho$. Many improvements on this method, which deal with the issues and complications affecting the standard algorithm, such as the updating process, initialization, and approximation of the learning rate, can be found in the literature. We refer the readers to the following papers: Power and Schoonees[177] explain in detail the standard mixture of Gaussians (MoG) method used by Stauffer and Grimson, Bouwmans *et al*.[59] discuss the improvements made to the standard MoG method, Carminati and Benois-Pineau[178] use an ISODATA algorithm to estimate the number of K Gaussians for each pixel, for the matching test the authors use the likelihood maximization followed by Markov regularization instead of the approximation of MAP, Kim *et al.*[179] show that an indoor scene is much closer to a Laplace distribution than to a Gaussian, for which a generalized Gaussian distribution (G-GMM) is proposed instead of a GMM to model the background. Makantasis *et.al*[180] propose to use the Student-t mixture model (STMM) rather than the Gaussian, due to the smaller number of parameters to be tuned. However, the use of STMM increases the complexity of calculation and the memory requirements. To solve this problem, the authors used an image grid, if change is detected using frame difference, the background modelling is applied in the corresponding grid. Many other works tried to use different mixture models like the Dirichlet[129] or hybrid (KDE-GMM) mixture model[78,181].

### *3.9 Spatio-Temporal Entropy Image*

In this method, which was proposed by Ma and Zhang[165], a statistical approach is adopted to measure the variation of each pixel based on its $w \times w$ neighbors along $L$ accumulated frames. A spatio-temporal histogram is created for each pixel, $H_{x,y,q}$ (Fig. 5), where $q$ denotes the bins of the histogram and the components of the histogram are $\{H_{x,y,1},...,H_{x,y,Q}\}$, where $Q$ is the total number of bins. Then, the corresponding probability density function for each pixel is given by[166]

$$P_{x,y,q} = \frac{H_{x,y,q}}{N} \tag{47}$$

Where $N = L \times w \times w$ .

To determine whether this pixel belongs to the background or foreground, an entropy measure, $E_{x,y}$, is computed from the probability density function.

$$E_{x,y} = -\sum_{q=1}^{Q} P_{x,y,q} \log\left(P_{x,y,q}\right) \tag{48}$$

Entropy is a measure of disorder; it is thus assumed that the entropy for a change due to noise is small compared to that due to a moving object.

The diversity of the state of each pixel indicates the intensity of motion at its position[165].

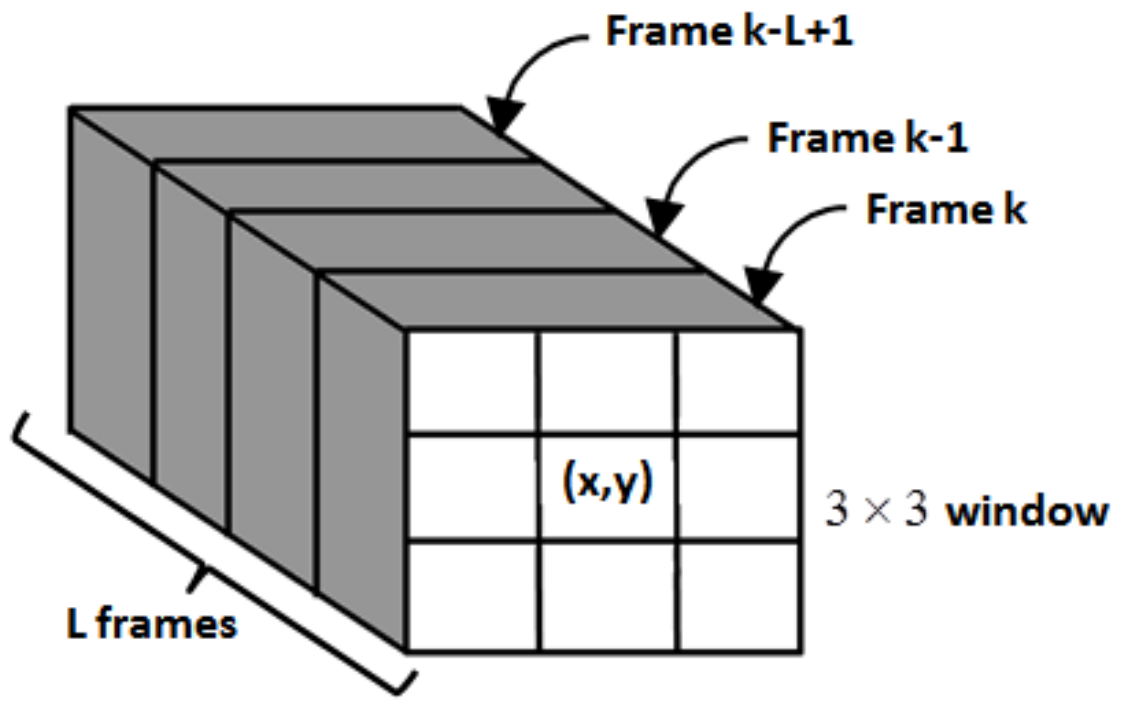

**Fig. 5.** Pixels used to construct the spatio-temporal histogram of pixel (i,j)

Ma and Zhang[165] first binarized the entropy result using an adaptive threshold and then applied a morphological filter (close-open operation) to enhance the results.

*3.10 Difference-based Spatio-Temporal Entropy Image (DSTEI)*

To overcome the problems associated with spatio-temporal entropy, especially the errors resulting from edge pixels, Jing *et al.*[166] attempted to use simple temporal differencing as follows:

$$D_t = \Phi\left(\left|I_t - I_{t-1}\right|\right) \tag{49}$$

where $\Phi$ quantizes the 256 grey-level values into $Q$ grey levels. As in the case of the STEI method, a spatio-temporal histogram is constructed for each pixel using a $w \times w$ window along a frame difference of $L$ as follows:

$$H_{x,y,q}(L) = \frac{1}{L} \sum_{k=1}^{L} h_{x,y,q}(k) \tag{50}$$

where $h_{x,y,q}(k)$ is the spatial histogram of each pixel in frame $k$.

To reduce memory consumption in a real-time system, the authors proposed recursive computation of the spatio-temporal histogram using (51), where $\alpha$ is a time constant that determines the influence of the previous frames:

$$H_{x,y,q}(k+1) = \alpha H_{x,y,q}(k) + (1-\alpha) h_{x,y,q}(k+1) \tag{51}$$

Then, the pdf of each pixel $P_{x,y,q}$ is obtained by normalizing (47). Subsequently, the entropy of each pixel is obtained using (48). Finally, a thresholding method is used to extract the motion region.

### *3.11 Eigen-background Subtraction*

The eigen-background method, or subspace learning using principle component learning (SL-PCA), is a background modelling method developed by Olivier *et al.*[42]. The concept underlying this method is that the moving object is rarely found in the same position in the scene across the training frames; hence, its contribution to the eigen-space model is not significant. Conversely, the static objects in the scene can be well described as the sum of various eigen-basis vectors. In this method, an eigen-space is formed using *N* reshaped training frames, $ES = [I_1\ I_2\ \ldots\ I_N]$, with mean $\mu$ and covariance $C$ .

$$\mu = \sum_{k=1}^{N} I_k \tag{52}$$

$$C = Cov(ES) = ES \cdot ES^T = \frac{1}{N}\sum_{k=1}^{N}[I_k - \mu]\cdot[I_k - \mu]^T \tag{53}$$

Then, we compute *M* principal eigenvectors by PCA, using singular value decomposition or eigen-decomposition; the eigen-decomposition is given by

$$C = V \Lambda V^T \tag{54}$$

where $\Lambda = diag\{\lambda_1, \lambda_2, ..., \lambda_N\}$ is the diagonal matrix that contains the eigenvalues ($\lambda_1 > \lambda_2 > ... > \lambda_N$) and $V$ is the eigenvector matrix. Further, $V_M$ consists of the $M$ eigenvectors in $V$ that correspond to the $M$ largest eigenvalues[169,182]. Then, every new image $I$ is normalized to the mean of the eigen-space $\mu$ and projected on these $M$ eigenvectors as follows:

$$I' = V_M^T (I - \mu) \tag{55}$$

Subsequently, the background is reconstructed by back-projection as follows:

$$B = V_M I' + \mu \tag{56}$$

Finally, the foreground can be detected by thresholding the absolute difference as

$$|I - B| > Th \tag{57}$$

The authors were very satisfied with the accuracy of the results obtained and specified that their method entails a lower computational load than the mixture of Gaussians method. However, they did not explain how to choose the images that form the eigen-space, because the model is based on the content of these images. If a scene includes a slow or large moving object or crowd movement in the eigen-space, the background model, which should contain only the static objects, will be inappropriate. Furthermore, the authors have not explained how to update the background to consider the possibility of moving objects becoming static. Many methods have been proposed to overcome these problems, e.g. updating the eigen-background[43,182] and using a selective mechanism[169]. A complete survey on PCA techniques applied to background subtraction can be found in the work of Bouwmans and Zahzah[69,81].

*3.12 Simplified Self-organized Background Subtraction (simplified SOBS)*

The use of a self-organizing map for background modelling was first proposed by Maddalena and Petrosino[21]. They built a model by mapping each color pixel $I_t(x, y)$ into an $n \times n$ weight vector, thus obtaining a neuronal map $B_t$ of size $\left[n \times W, n \times H\right]$, where $W$ and $H$ are the width and height of the observed scene. The initial neuronal map $B_0$ is obtained in the same manner, where $I_0$ represents the scene containing the static objects (Fig. 6).

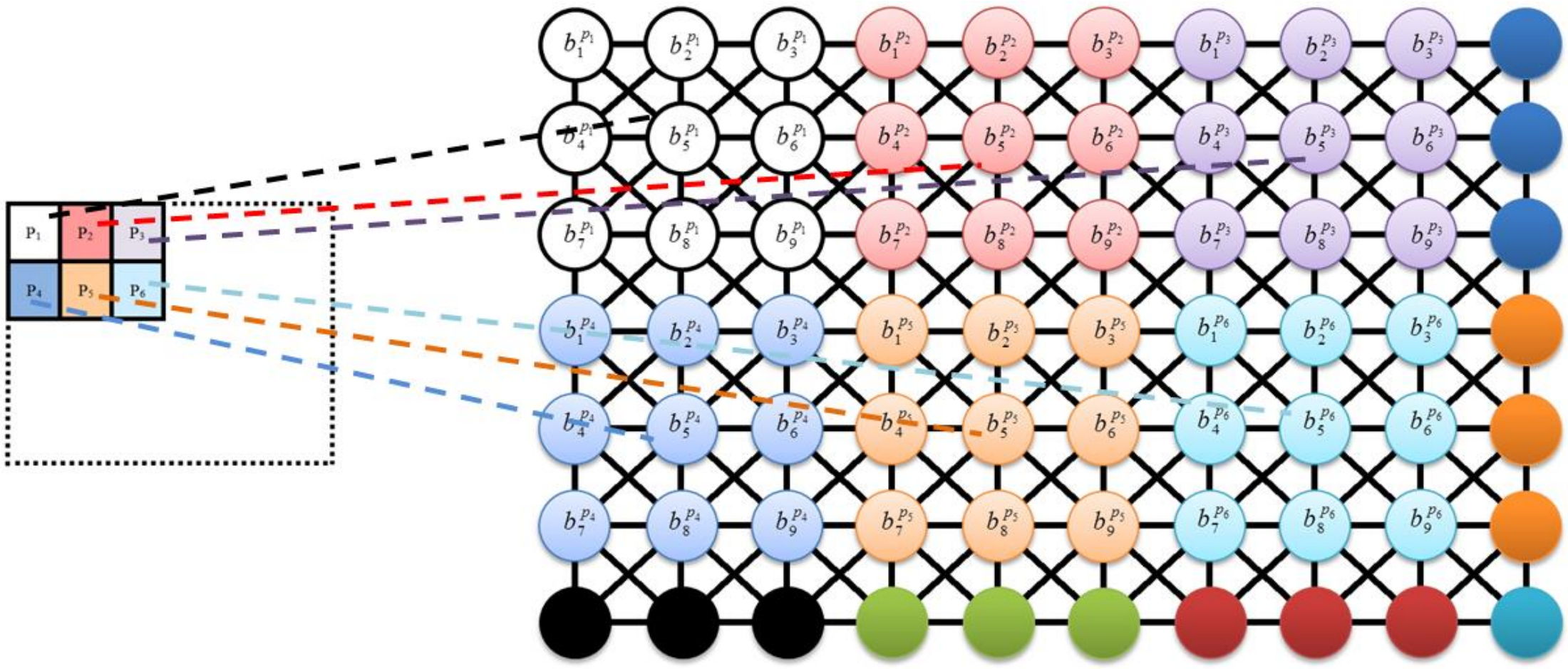

**Fig. 6.** (Left) A simple image $I_t$; (Right) the modelling neuronal map $B_t$ when n = 3

Then, for each pixel $I_t\left(x, y\right)$ from the incoming frame, a matching test is performed with the corresponding weights $b_i, i = 1, ..., n^2$ to find the best match $b_m$ defined by the minimum distance between $I_t(x, y)$ and $b_i$ that should not be greater than a predefined threshold $Th$

$$d_{\min}\left(b_m, I_t(x,y)\right) = \min_{i=1,...,n^2}\left\{d\left(b_i, I_t(x, y)\right)\right\} \leq Th \tag{58}$$

The difference is computed per the color space of the image. Maddalena and Petrosino[21] used the HSV hexagonal cone color space; the Euclidean distance in this case is given by[172]

$$d\left(b_i, I_t(x,y)\right) = \sqrt{\left(\upsilon_{b_i} s_{b_i} \cos\left(h_{b_i}\right) - \upsilon_{I_t} s_{I_t} \cos\left(h_{I_t}\right)\right)^2 + \left(\upsilon_{b_i} s_{b_i} \sin\left(h_{b_i}\right) - \upsilon_{I_t} s_{I_t} \sin\left(h_{I_t}\right)\right)^2 + \left(\upsilon_{b_i} - \upsilon_{I_t}\right)^2} \tag{59}$$

If the best match is found in background $B_t$ at position $\left(\bar{x}, \bar{y}\right)$, we consider the pixel $I_t\left(x, y\right)$ to be a background pixel; otherwise, we consider it to be a foreground pixel. We update the model around the best match position as follows:

$$b_{t+1}(i, j) = b_t(i, j) + \alpha_{i,j}(t)\left(I_t(x, y) - b_t(i, j)\right) \tag{60}$$

For i=x̄-$\lfloor n/2 \rfloor$,...,x̄+$\lfloor n/2 \rfloor$, j=ȳ-$\lfloor n/2 \rfloor$,...,ȳ+$\lfloor n/2 \rfloor$, $\alpha_{i,j}(t) = \alpha(\mathrm{t})\omega_{i,j}$ , where $\omega_{i,j}$ are $n \times n$ Gaussian weights and $\alpha(\mathrm{t})$ is the learning factor given by

$$\alpha(\mathrm{t}) = \begin{cases} \alpha_1 - t\left(\dfrac{\alpha_1 - \alpha_2}{K}\right), & if \quad 0 \le t \le K \\ \alpha_2, & if \quad t > K \end{cases} \tag{61}$$

where $\alpha_1$ and $\alpha_2$ are predefined constants such that $\alpha_2 \le \alpha_1$ and $K$ is the number of frames required for the calibration phase, which depends on how many static initial frames are available for each sequence.

To reduce the computational load as well as the number of parameters to be tuned, Chacon *et.al*[24] proposed an SOM-like architecture in which the mapping is one-to-one, i.e. each neuron is associated with its corresponding pixel. Since each pixel $I_t(x, y)$ is represented in the HSV color space, each neuron $b(x, y)$ has three inputs $h_b, s_b, \upsilon_b$ , and the matching test is performed as follows:

$$d\left(b, I_t(\mathrm{x,y})\right) \le Th \quad \wedge \quad \left|\upsilon_{I_t} - \upsilon_b\right| \le Th_v \tag{62}$$

where $d\left(b, I_t(\mathrm{x,y})\right)$ is the Euclidean distance in HSV color space, defined previously in (59), and $Th$ and $Th_v$ are threshold values experimentally set by the authors. The second condition eliminates object shadows. If the result is true, the current pixel is considered to be a background pixel and the weights of the corresponding neuron $b(x, y)$ and its neighbors are updated using (63) and (64), respectively.

$$b_{t+1}(x, y) = b_t(x, y) + \alpha_1\left(I_t(x, y) - b_t(\mathrm{x}, y)\right) \tag{63}$$

$$b_{t+1}(x', y') = b_t(x', y') + \alpha_2\left(I_t(x', y') - b_t(x', y')\right) \tag{64}$$

where $x' = x$ -1, x+1 and $y' = y$ -1, y+1 are the coordinates of the neighboring neurons, and $\alpha_1$ and $\alpha_2$ are the learning rates, with $\alpha_1 > \alpha_2$ for non-uniform learning.

If the result of (62) is not true, the current pixel is considered to be a foreground pixel and no update is required. The authors showed that this simplified model performs satisfactorily in different scenarios.

Many other studies have attempted to improve the original self-organized background subtraction method (SOBS). For example, Maddalena and Petrosino[65] introduced fuzzy rules for subtraction and process updates. Further, the authors[183,184] used a 3D neuronal map to model the background; the map consists of *n* layers of the classical 2-grid neuronal map and considers scene changes over time.

## 4 Evaluation Metrics and Performance Analysis

To correctly evaluate these methods and achieve a fair comparison, the methods were applied to the same dataset. Further, to define the properties of each method, we used the same seven metrics as those used for CDnet2014: recall, specificity, false positive rate (FPR), false negative rate (FNR), percentage of wrong classification (PWC), precision, and F-measure. The role of these metrics is to quantify how well each algorithm matches the ground truth. All metrics are based on the following four quantities[79]:

True positives (TP): number of foreground pixels correctly detected

False positives (FP): number of background pixels incorrectly detected as foreground pixels

True negatives (TN): number of background pixels correctly detected

False negatives (FN): number of foreground pixels incorrectly detected as background pixels (also known as misses)

Recall (the sensitivity or true positive rate) is the ratio of the number of foreground pixels correctly detected by the algorithm to the number of foreground pixels in the ground truth[79].

$$\mathrm{Re} = \frac{TP}{TP + FN} \tag{65}$$

Specificity (the true negative rate) represents the percentage of correctly classified background pixels.

$$Sp = \frac{TN}{TN + FP} \tag{66}$$

False Positive Rate (FPR) is the ratio of the number of background pixels incorrectly detected as foreground pixels by the algorithm to the number of background pixels in the ground truth.

$$FPR = \frac{FP}{TN + FP} \tag{67}$$

False Negative Rate (FNR) is the ratio of the number of foreground pixels incorrectly detected as background pixels by the algorithm to the number of background pixels in the ground truth.

$$FNR = \frac{FN}{FN + TP} \tag{68}$$

Percentage of Wrong Classification (PWC) is defined as the percentage of wrongly classified pixels.

$$PWC = \frac{FN + FP}{TN + TP + FP + FN} \tag{69}$$

Precision is defined as the ratio of the number of foreground pixels correctly detected by the algorithm to the total number of foreground pixels detected by the algorithm[79,94]

$$\mathrm{Pr} = \frac{TP}{TP + FP} \tag{70}$$

F-measure (or F1 score) is a measure of quality that quantifies, in one scalar value for a frame, the similarity between the resulting foreground detection image and the ground truth[100,59]. Mathematically, it is a trade-off between recall and precision[185]

$$F1 = 2.\frac{\mathrm{Pr}.\mathrm{Re}}{\mathrm{Pr} + \mathrm{Re}} \tag{71}$$

For comparison, we have adopted the same approach used on CDnet[109,116] to generate results. First, for each method, we computed all the metrics for each video in each category; then a category average metric was computed:

$$M_{\mathrm{c}} = \frac{1}{N_c} \sum_{v} M_{v,c} \tag{72}$$

where $M_{\mathrm{c}}$ represents one of the seven metrics (Re, Sp, FPR, FNR, PWC, Pr, F1), $N_c$ is the number of videos in each category, and *v* is a video in category *c*.

We also defined an overall average metric (OAM), which is the simple average of the category averages:

$$OAM = \frac{1}{C} \sum_{c=1}^{C} M_c \tag{73}$$

where *C* is the number of categories.

To rank all the methods, for each category *c*, we computed the rank of each method for metric *M*. Then, we computed the average rank of this method across all the metrics:

$$RM_c = \frac{1}{7} \sum_{n=1}^{7} Rank(M_c) \tag{74}$$

Subsequently, we computed the average over all categories to obtain the average rank across categories *RC* for each method:

$$RC = \frac{1}{C} \sum_{c=1}^{C} RM_c \tag{75}$$

In addition, we computed the average rank across the OAM for each method:

$$R = \frac{1}{7} \sum_{n=1}^{7} rank\,(\mathrm{OAM}) \tag{76}$$

# 5 Results and Discussion

We applied each motion detection method described in Section 2 to the CDnet 2014 dataset[125], which includes different scenarios and challenges. Fig. 7 shows sample frames from each video in each category, while Fig. 8 shows their corresponding ground truths.

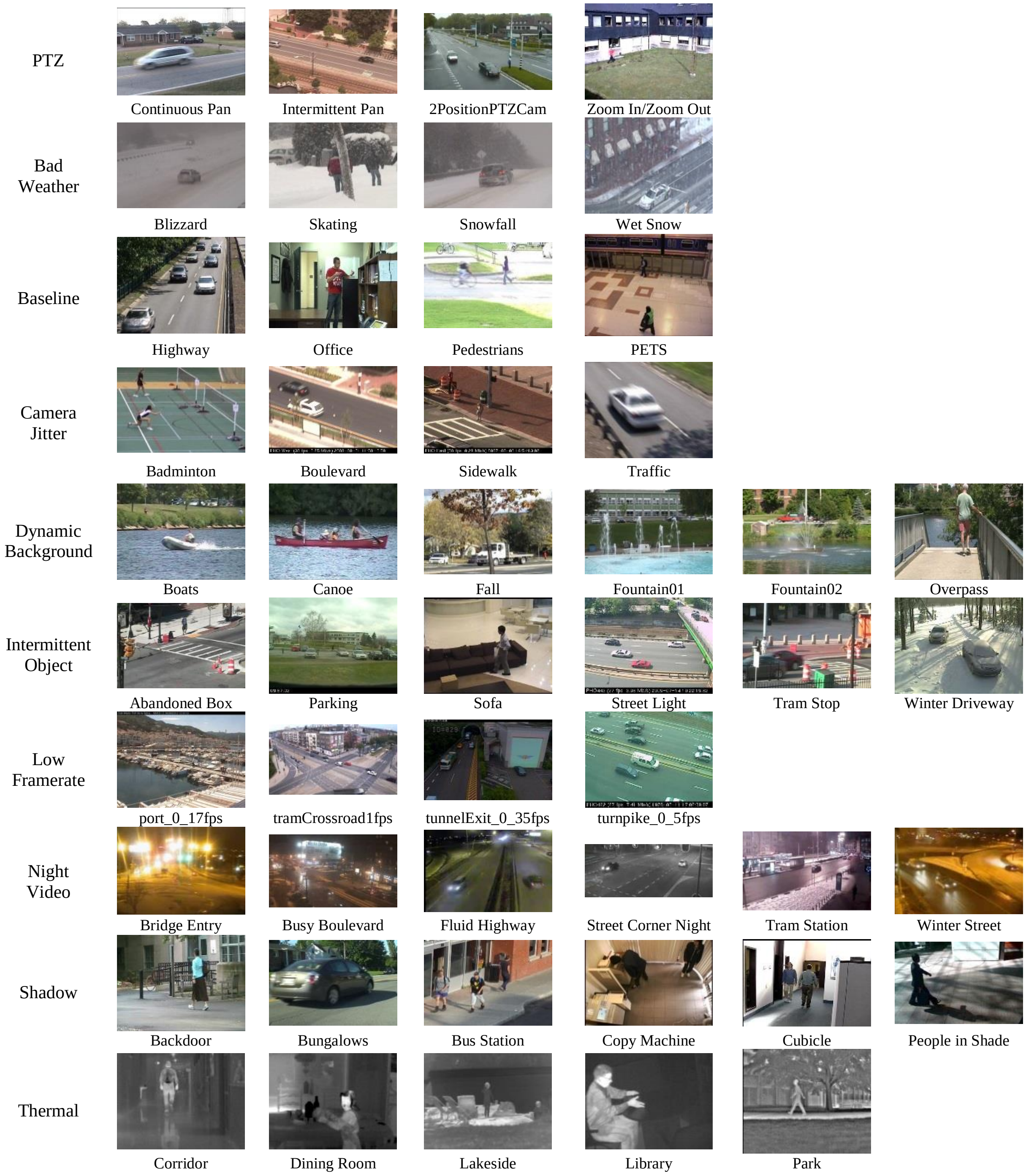

Turbulence

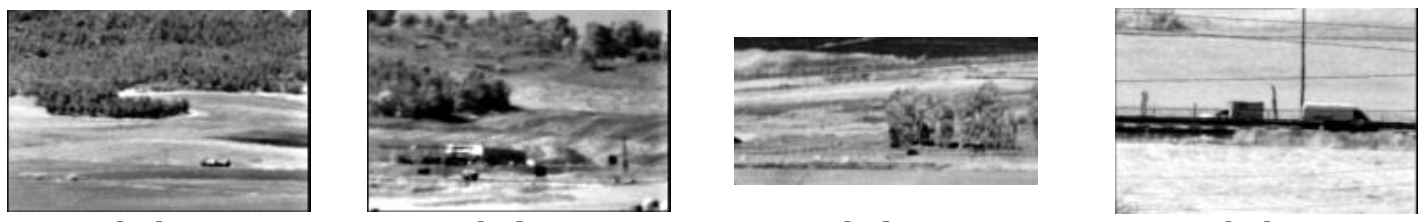

Turbulence0 Turbulence1 Turbulence2 Turbulence3

**Fig. 7.** Sample frames from each video in each category

Each motion detection method was followed by an automatic thresholding operation in order to determine region changes and remove small changes in luminosity, except for the RGA, MoG, and MRFMD methods; for the two first methods, the threshold was fixed to $2.5\sigma$, where $\sigma$ denotes the standard deviation, and for MRFMD, a fixed threshold $Th = 35$ was used to compute the observation $O(x, y, t)$. We selected Otsu's thresholding method based on a previous study[67]. For the eigen-background method, we set the number of training images to $N$ = 28. These training images were equally spaced by 10 frames and the number of eigen-background vectors was set to $M$ = 3.

For the MoG method, the parameters used were selected in accordance with the work of Nikolov *et al.*[185], who measured the accuracy of the algorithm as a function of each variable parameter. Further, they proposed a set of optimal parameters to improve the performance of the MoG algorithm. Accordingly, we selected the number of Gaussians as $K$ = 3, the learning rate as $\alpha = 0.01$, the foreground threshold as $T = 0.25$, the deviation threshold as $D = 2.5$, and the initial standard deviation as $\sigma_{init} = 20$.Notice that the selected parameters are different from those presented in CDnet2014[125], furthermore, we adopted the approximation of Power and Schoonees[177] to compute the learning rate $\rho$, hence the values of $(\mu_t, \sigma_t)$ were affected and different too.

For the running Gaussian average method, the learning rate was set to $\alpha = 0.01$ and the deviation threshold was set to $D = 2.5$. We note that MoG and RGA were applied to grey-level images in all our tests. We also applied the STEI and DSTEI methods to the grey-level images; to construct the spatio-temporal histogram, we selected a 3×3 window with 5 images and the number of grey levels was set as $Q$ = 100.

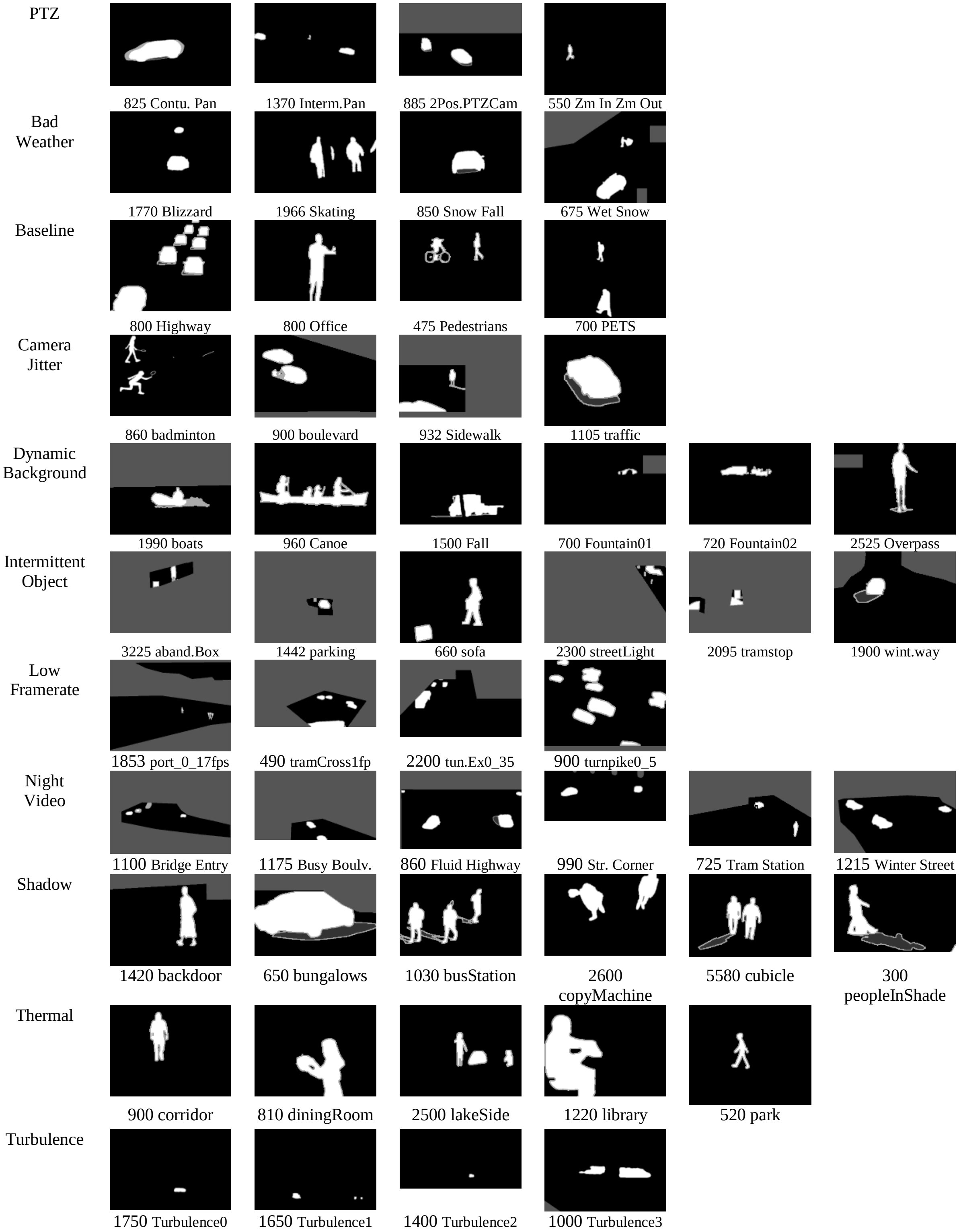

**Fig. 8.** Corresponding ground truths of the sample frames in Fig. 7

For the MRF-based motion detection algorithm, per the work of Caplier[160], we set the four parameters to $\beta_s = 20, \beta_p = 10, \beta_f = 30, \alpha = 10$. For the $\Sigma\Delta$ method, the only parameter to be set was *N*; we selected *N* = 3 (typically, *N* = 2, 3, or 4)[158]. For the simplified self-organized background subtraction, the method was tested on the HSV color space, where four parameters, $Th, Th_v, \alpha_1$ and $\alpha_2$, had to be set; $Th$ and $Th_v$ were set automatically using Otsu's method, and the learning rates were set to $\alpha_1$=0.02 and $\alpha_2 = 0.01$, according to Ref.[60]. Further, a median filter with *L* = 10 frames was used to initialize the weights of the neuronal map $B_0$. Finally, for the forgetting morphological temporal gradient method and the adaptive background detection method, the parameter $\alpha$ should take values in the interval $[0,1]$. In our tests for these last two methods, we chose $\alpha = 0.1$. For FD and 3FD, we applied these methods on grayscale images by using Eq. (1) and Eq. (5-8) respectively.

Table 2 summarizes the selected parameters for each tested method.

**Table 2** Selected parameters for each method

| Method | Abbrev. | Parameters used |
|---|---|---|
| Temporal differencing[7,86] | FD | N/A |
| Three-frame difference[151,152] | 3 FD | N/A |
| Running average filter[90,154] | RAF | $\alpha = 0.1$ |
| Forgetting morphological temporal gradient[156] | FMTG | $\alpha = 0.1$ |
| $\Sigma\Delta$ background estimation[157] | $\Sigma\Delta$ | N = 3 |
| MRF-based motion detection algorithm[159] | MRFMD | $\beta_s = 20, \beta_p = 10, \beta_f = 30, \alpha = 10$ |
| Spatio-temporal entropy image[165] | STEI | w×w×L = 3×3×5, Q = 100 |
| Difference-based spatio-temporal entropy image[166] | DSTEI | w×w×L = 3×3×5, Q = 100 |
| Running Gaussian average[14] | RGA | $\alpha = 0.01, D = 2.5$ |
| Mixture of Gaussians[15] | MoG | $K = 3, \alpha = 0.01, T = 0.25, D = 2.5$ |
| Eigen-background[42] | Eig-Bg | N = 28, M = 3 |
| Simplified self-organized background subtraction[24] | Simp-SOBS | $\alpha_1$=0.02 and $\alpha_2 = 0.01$ |

The overall results of testing these methods using the CDnet dataset (CDnet2012 and CD2014) are reported in Table 3, where entries are sorted by their average rank across categories (RC).

**Table 3** Overall results across all categories

| | Recall | Specificity | FPR | FNR | PWC | Precision | F-Measure | R | R C |
|---|---|---|---|---|---|---|---|---|---|
| Simp-SOBS | 0,49362 | 0,97220 | 0,02780 | 0,50638 | 4,67079 | 0,51477 | 0,40097 | 4,00000 | **4,68831** |
| RGA | 0,30123 | 0,99351 | 0,00649 | 0,69877 | 3,22117 | 0,49415 | 0,31465 | **3,42857** | 5,15584 |
| GMM | 0,20606 | **0,99593** | 0,00407 | 0,79394 | **3,08499** | **0,61021** | 0,25420 | 4,00000 | 5,44156 |
| Eig-Bg | **0,59669** | 0,93715 | 0,06285 | **0,40331** | 7,35814 | 0,41815 | **0,41028** | 6,57143 | 6,00000 |
| RAF | 0,36107 | 0,97060 | 0,02940 | 0,63893 | 5,25655 | 0,44158 | 0,27924 | 6,28571 | 6,05195 |
| FMTG | 0,42449 | 0,95736 | 0,04264 | 0,57551 | 6,37002 | 0,42101 | 0,28152 | 7,14286 | 6,20779 |
| DSTEI | 0,29669 | 0,96815 | 0,03185 | 0,70331 | 5,63380 | 0,41881 | 0,22299 | 8,14286 | 6,67532 |
| FD | 0,22779 | 0,97247 | 0,02753 | 0,77221 | 5,43072 | 0,46649 | 0,18825 | 6,85714 | 6,83117 |
| 3FD | 0,08117 | 0,98815 | 0,01185 | 0,91883 | 4,27114 | 0,46440 | 0,08201 | 8,00000 | 6,94805 |
| MRFMD | 0,08693 | 0,99056 | **0,00944** | 0,91307 | 4,02845 | 0,42689 | 0,09293 | 7,00000 | 7,37662 |
| $\sum\Delta$ | 0,13762 | 0,98851 | 0,01149 | 0,86238 | 4,11532 | 0,37271 | 0,14229 | 7,42857 | 7,49351 |
| STEI | 0,45870 | 0,78646 | 0,21354 | 0,54130 | 22,18321 | 0,12255 | 0,12881 | 9,14286 | 9,12987 |

It is clear from Table 3 that the STEI method generated poor results compared to the other methods; the use of entropy alone as a metric to detect moving objects did not yield good results because the spatio-temporal accumulation window may contain object edges, which can lead to high diversity (high entropy) and thus impair the segmentation result. Moreover, this error can spread to the entire edge region (see Fig.16 and Fig.17), generating a very high PWC, high FPR and very low percentage of correctly classified background pixels (Sp). STEI is also very sensitive (high recall) due to low misses (False negatives, see Fig.9).

Adding the frame difference to this method (DSTEI) increased its precision and decreased the PWC considerably (Fig.14, Fig.13), except for the "Camera Jitter" category which still has high FPR, PWC and low F-measure (Fig.11, 13, 15). Moreover, from the overall results in Table 3, we note that DSTEI did not achieve significant improvement over the FD method, owing to the drawbacks of using the spatio-temporal accumulation window and the tails caused by using inappropriate values of $\alpha$ to compute the spatio-temporal histogram recursively (see Fig.16 and Fig.17). Notably, DSTEI has acceptable percentage of correctly classified background pixels (Sp) in the "Dynamic Background" category, compared to other methods, see Fig.10.

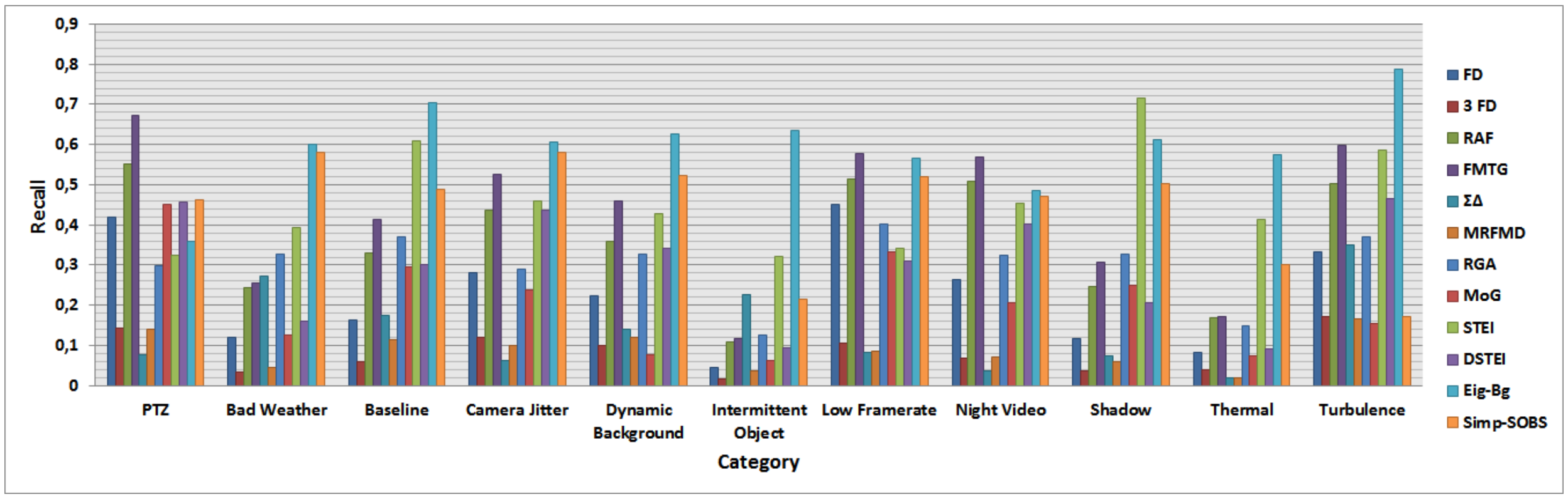

**Fig. 9.** Recall results for all tested methods over all categories

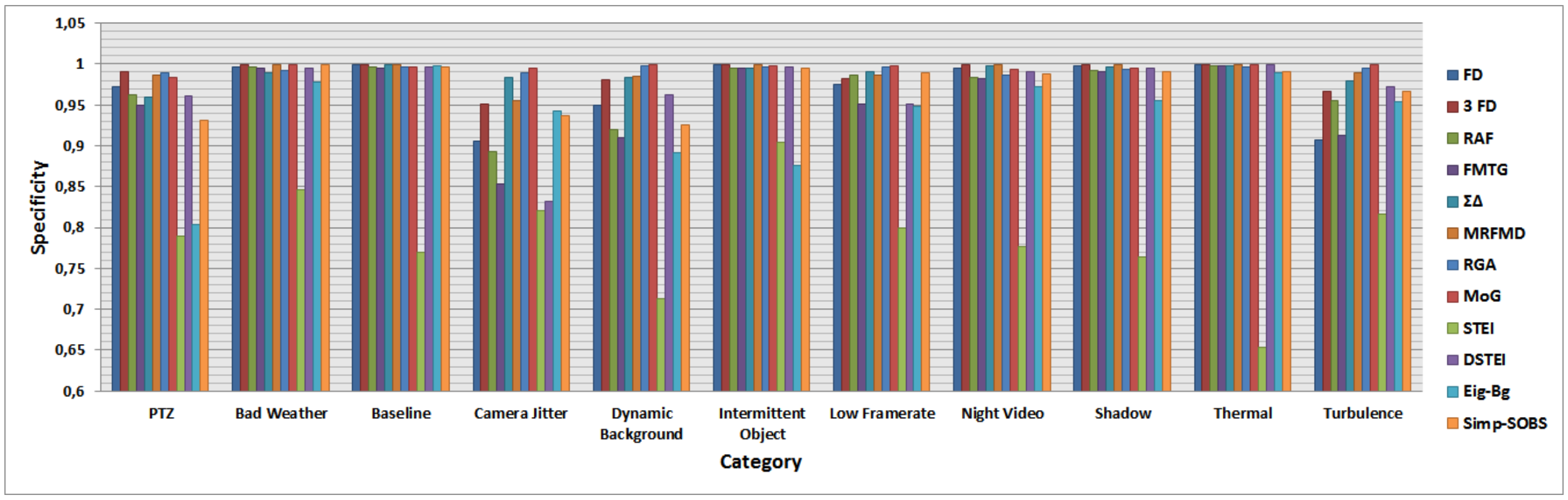

**Fig. 10.** Specificity results for all tested methods over all categories

From Table 3, we can see that the $\Sigma\Delta$ method produced poor results but achieved significant improvement over the STEI method. The $\Sigma\Delta$ method was characterized by a low FPR (Fig.11) and high specificity (Fig.10), i.e. many background pixels were correctly classified, but it still suffered from a high false negative rate, especially in "PTZ", "Camera Jitter" and "Thermal" categories, and also low precision in "PTZ", "Bad Weather", "Dynamic Background", "Shadow" and "Thermal" categories. In, Fig. 16, we observe that false detections caused by snowfall in the 'Skating' video were eliminated, whereas the objects in motion were not well segmented. We note also that this method has a high precision in the case of "Baseline" category, Fig.14, however, results on this category, Fig.15, shows holes left in the segmented objects (high misses) which produce a low recall (Fig.9), this problem is owed to the initialization step based on temporal differencing.

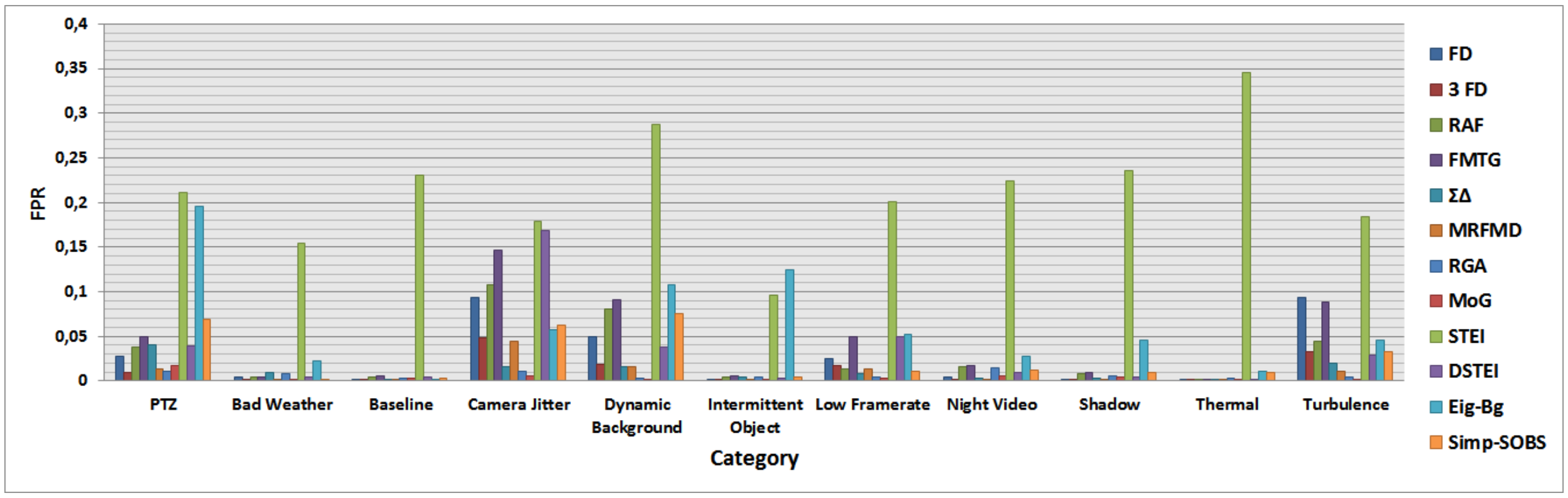

**Fig. 11.** False positive rate results for all tested methods over all categories

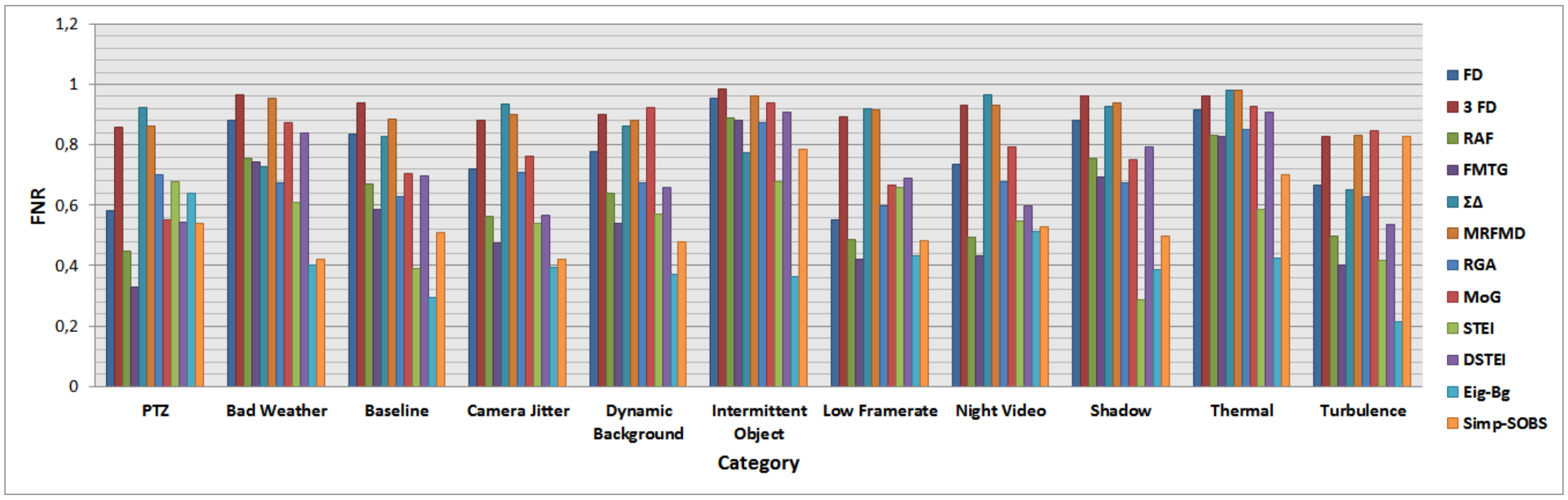

**Fig.12.** False negative rate results for all tested methods over all categories

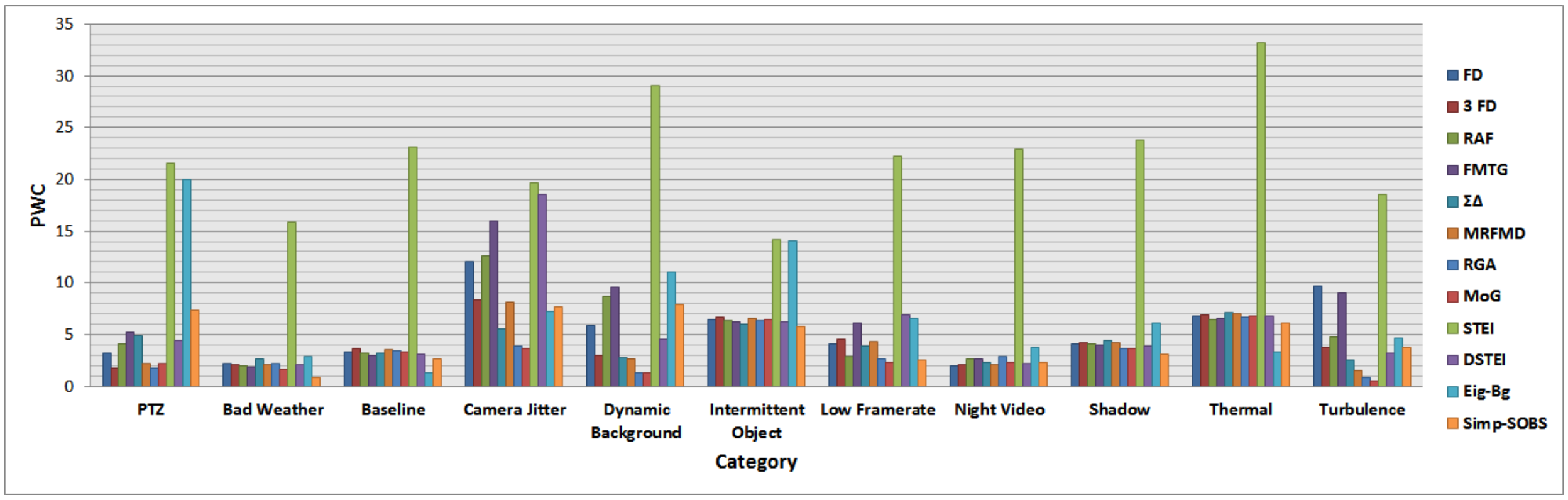

**Fig.13.** Percentage of wrong classification results for all tested methods over all categories

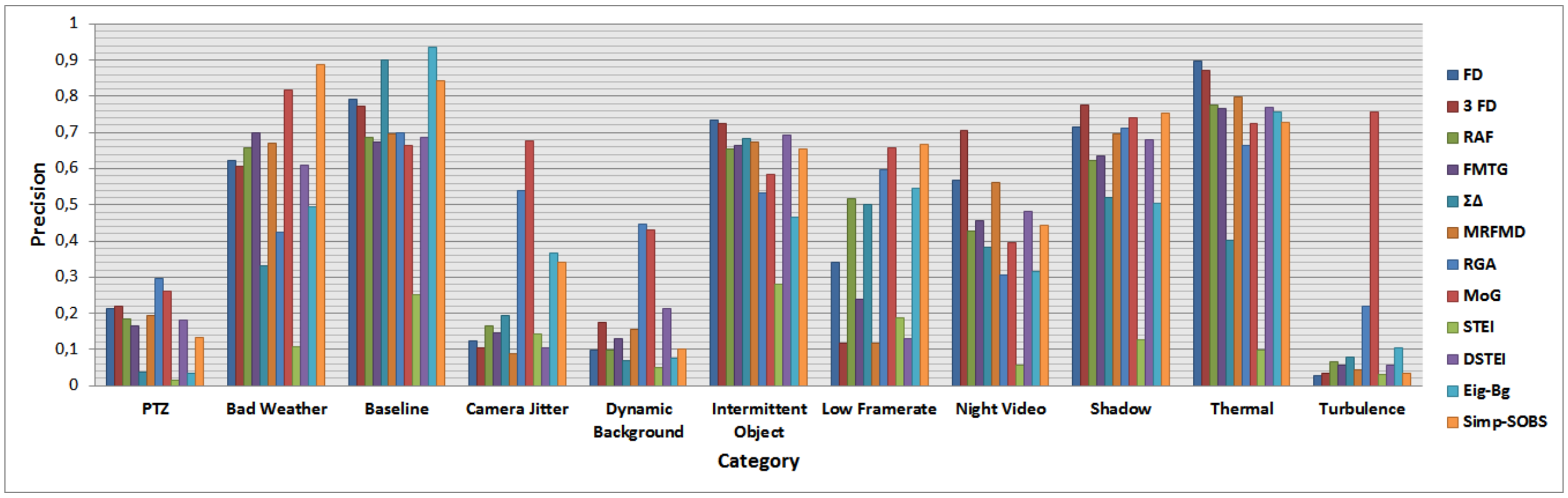

**Fig.14.** Precision results for all tested methods over all categories

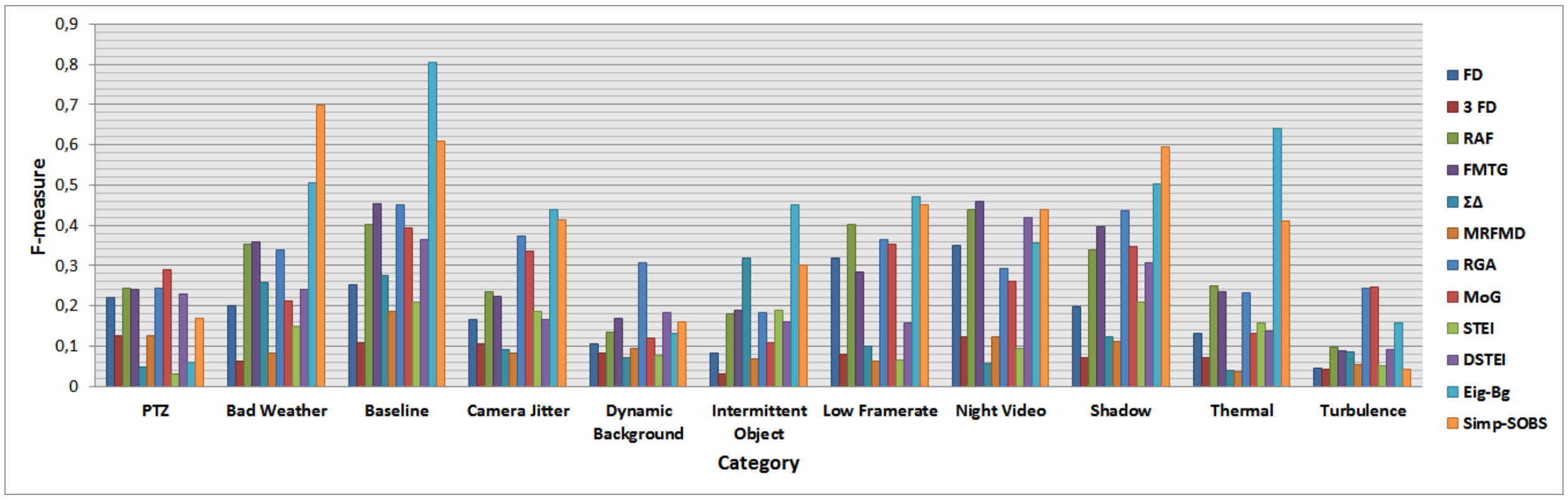

**Fig.15.** F-measure results for all tested methods over all categories

The MRFMD method also yielded poor results, and was similar to the $\Sigma\Delta$ method, with a high FNR and low Recall (Fig.12, Fig.9), due to its dependence on the initialization step based on the temporal differencing method, leading to incompletely segmented moving objects (holes). Nevertheless, this method performed well by enhancing the image difference in terms of specificity (Fig.10) , PWC (Fig.13) and has acceptable precision in the cases of "Bad Weather", "Shadow", "Night Video", and "Thermal" (see Fig.14 & Appendix, Table 2).

In addition, this method is parametric because we had to define the values of the model energy ($\beta_s, \beta_p, \beta_f, \alpha$). Furthermore, the iterative conditional mode (ICM) technique is a suboptimal algorithm that may converge to local minima, but its computational time is considerably shorter than that of a stochastic relaxation scheme (i.e. simulated annealing)[159].

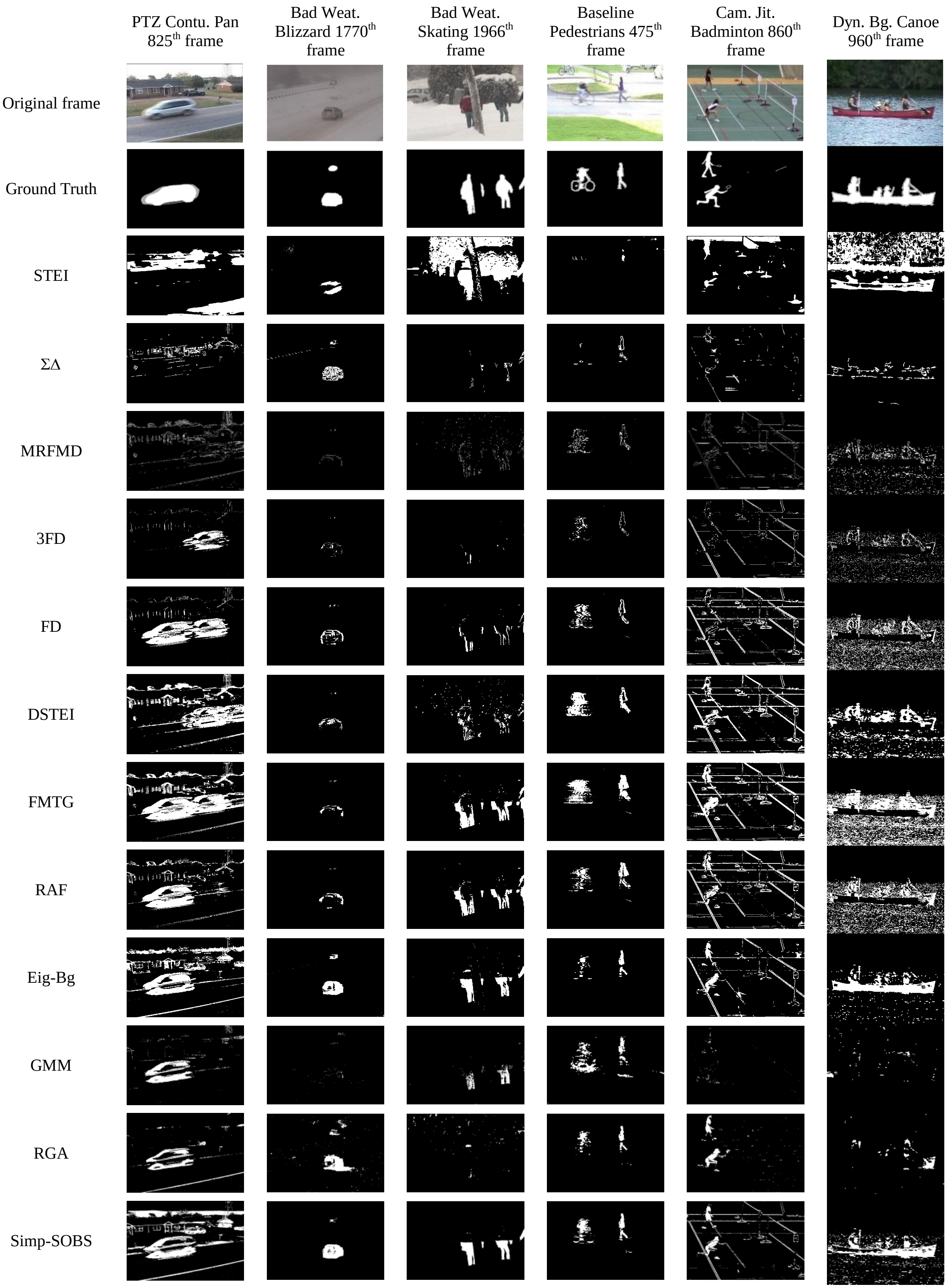

Fig. 16. Samples from the results of tested motion detection methods (PTZ, Bad Wea., Baseline,Cam Jit.,Dyn. Bg

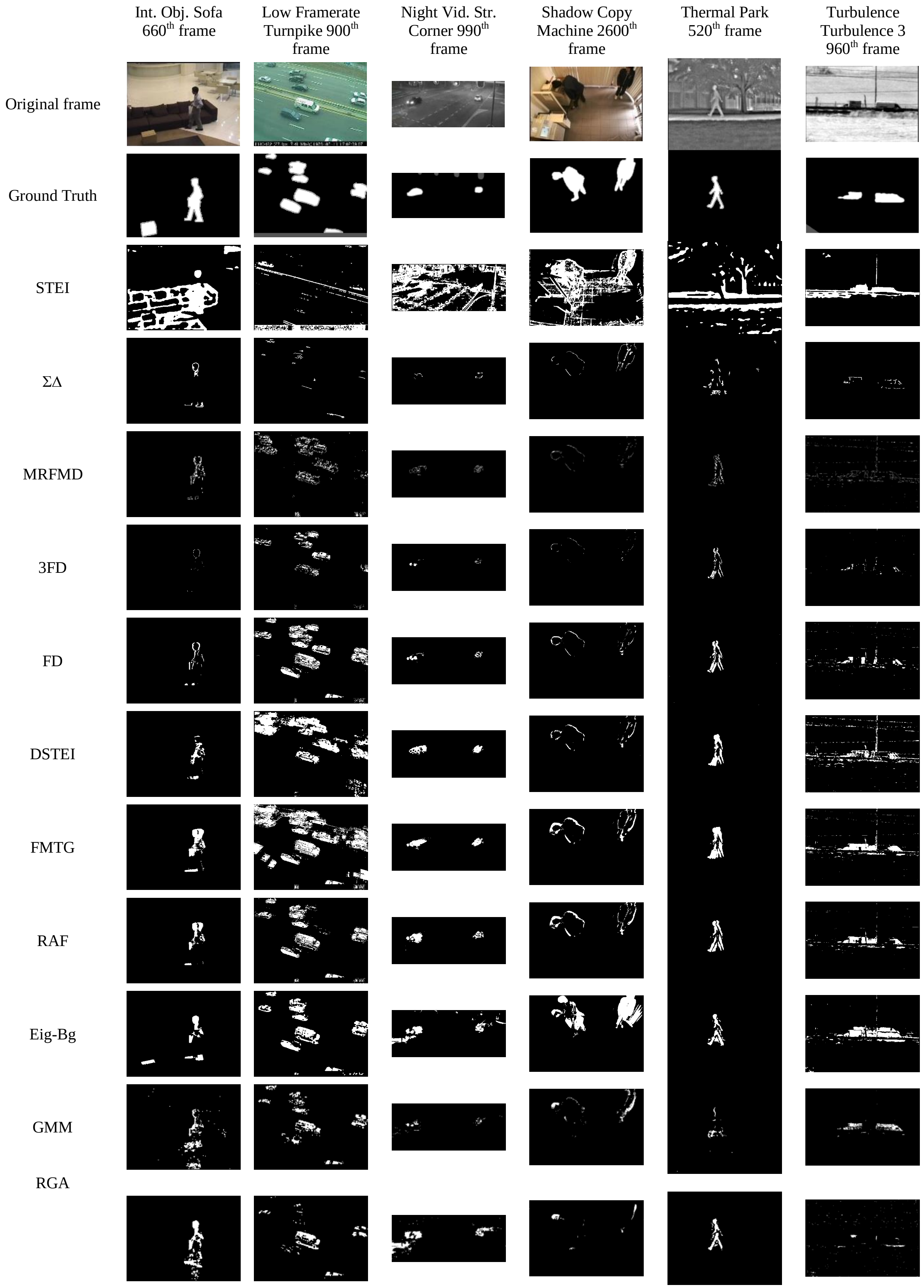

Int. Obj. Sofa 660th frame
Low Framerate Turnpike 900th frame
Night Vid. Str. Corner 990th frame
Shadow Copy Machine 2600th frame
Thermal Park 520th frame
Turbulence Turbulence 3 960th frame
Original frame
Ground Truth
STEI
ΣΔ
MRFMD
3FD
FD
DSTEI
FMTG
RAF
Eig-Bg
GMM
RGA

Simp-SOBS

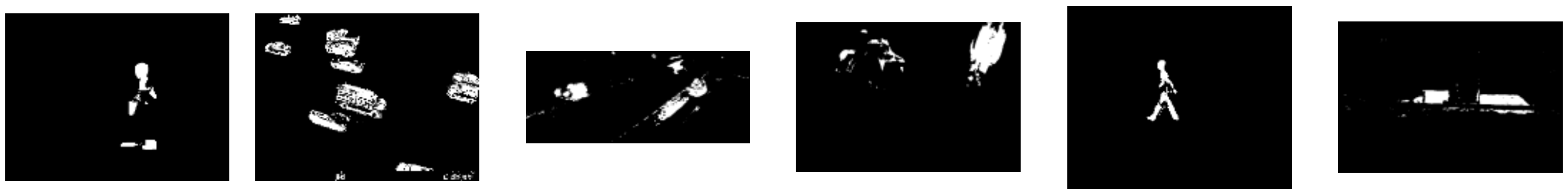

Fig. 17. Samples from the results of tested motion detection methods (Int.Obj., Low.Fr., N.Vid., Shad., Ther., Turb.)

As expected, the frame difference and three-frame difference methods did not show good results, except for high precision, mainly because of the incomplete segmentation of the shape of moving objects, preserving only the edges. Moreover, these methods suffered from overlap of slow moving objects and poor detection of objects far from the camera. To overcome the problem of incomplete segmentation, the threshold operation in the frame difference method is usually followed by morphological operations to link the edges of the moving objects. Then, regions and holes in the image are filled. Another solution is to combine frame difference methods with a background subtraction method[5,186]. We also note that these methods demonstrate good detection of foreground pixels in night-time videos compared to background subtraction techniques, see Fig.14 and Appendix Table 8, owing to the change of light (vehicle or street light) which impairs the modelled background.

The forgetting morphological temporal gradient method (FMTG) yielded acceptable results with observable low FNR (Fig.12) and high recall (Fig.9) for “PTZ”, “Camera Jitter”, “Dynamic Background”, “Low Framerate”, “Night Video” and “Turbulence” categories. However, it was characterized by high FPR, caused by artificial tails due to an inappropriate value of $\alpha$. Moreover, this method was conducive to strong background motion, as in the “Dynamic Background” category, Fig16.

As with FMTG, the RAF method yielded acceptable results. This result was expected because FMTG is based on a recursive operation of the running average filter, i.e. Eq. (10). However, FMTG gave good detection results in bad weather conditions (see Fig.14, Fig.15 & Appendix, Table 2) owing to the morphological temporal gradient filter, Eq. (15). Compared to FD and 3FD, RAF method outperforms them for almost all categories (except night videos) in terms of F-measure, FNR and sensitivity, but with high FPR.

The fourth-best method according to the evaluation results (Table 3) was the eigen-background method, which yielded good detection results as well as silhouettes of objects. We noted that it had a high recall among all categories, Fig.9 (except “Low Framerate” category), and low FPR, Fig.11. For “Intermittent Object” and “PTZ” categories the Eig-Bg shows high FPR and PWC (Fig. 11 and Fig.13), this is due to the absence of update process in the original algorithm. Different techniques have been developed to resolve this problem, to this end we refer the reader to these Ref.[43,69,81,181]. Remarkably, this method had a great precision and F-measure for the “Baseline” category which makes this method the ideal approach for easy and mild challenging scenes. However, its results were strongly dependent on the images that form the eigen-space; the presence of moving objects in this space could alter the detection results. The execution time of this method depended on the number of eigenvectors. In our case, the execution time seemed acceptable for real-time application; however, memory requirements make it unsuitable for this type of applications.

Methods based on Gaussian distributions showed better performance than other methods, the GMM method is characterized by its very high precision and high F1-score in difficult challenging categories (“Bad weather, Dynamic Background, Camera Jitter, Low framerate, Shadow and Turbulence”), and has low PWCs and low FPR in all categories, this is due to the number of K Gaussians used to model the dynamic backgrounds. Notably, the RGA method outperforms the MoG method (Table 3), with higher Recall (Fig.9), low FNR values (Fig.12) and higher F-measure (Fig.15) in almost all categories. This is owed to the large number of parameters required to set for the MoG algorithm ($K, \alpha, T, D, \sigma$), which differ with the challenging conditions presented by a video (day/night, indoor/outdoor, complex/simple background, with/without noise). Thus, in some cases, the RGA method seemed to be sufficient. Moreover, its computational complexity was lower than that of the MoG method, as was found by Piccardi[4] and Benezeth *et al.*[97].

Finally, Simp-SOBS showed the best results of all the methods, with high precision, recall, and F-measure, owing to the use of the HSV color space and the condition on the $V$ value component that significantly reduced object shadows. Further, we note from Table 4 that the results of this method in challenging categories (‘Bad weather’, ‘Low frame rate’ and ‘Shadow’) were as good as those in simple categories (‘Baseline’). From the previous results, we can note that the most challenging categories for this method are: the “Turbulence” category characterized by low precision (Fig.14) and high FNR (Fig.12), and the “PTZ” category characterized by high PWC (Fig.13) and high FPR (Fig11).

**Table 4** Top three methods for all categories based on the average rank across categories (RC)

| | 1st | 2nd | 3rd |
|---|---|---|---|
| PTZ | MoG | RGA | RAF |
| Bad Weather | Simp-SOBS | MoG | FMTG |
| Baseline | Eig-Bg | Simp-SOBS | ΣΔ |
| Camera Jitter | Eig-Bg | RGA | MoG |
| Dynamic Background | RGA | DSTEI | MoG |
| Intermittent Object Motion | ΣΔ | Simp-SOBS | DSTEI |
| Low Framerate | Simp-SOBS | RGA | MoG |
| Night Video | FD | 3FD | DSTEI |
| Shadow | Simp-SOBS | RGA | MoG |
| Thermal | Eig-Bg | RAF | FD |
| Turbulence | RGA | MoG | Eig-Bg |

Figure 18 shows the frame rate (execution time) of each method, applied on “PETS2006” video from the “Baseline” category, with a resolution of $720 \times 576$. Tests were carried on Intel I7 2.3 Ghz with 16 GB RAM, parts of the code was non-vectorized. The Figure shows that the fastest methods were DF, 3FD, RAF and FMTG because of their simplicity, Eig-Bg shows also a good execution time, this is interpreted by the use of subset of singular values and vectors which overcomes the long time required to compute the N eigenvectors using Eigen decomposition. ΣΔ has a slow frame rate owing to the post processing step required to eliminate the ghost effect. The slowest methods were MoG, STEI and DSTEI; MOG is slow because of the computing complexity linked to the use of K Gaussian distributions, the time required to update their parameters, and to order them; and the slowness of STEI and DSTEI is more closely related to the time needed to compute the histogram from the spatio-temporal window. MRF, Simp-SOBS and RGA also had long execution times, but they were not as slow as the previous methods.

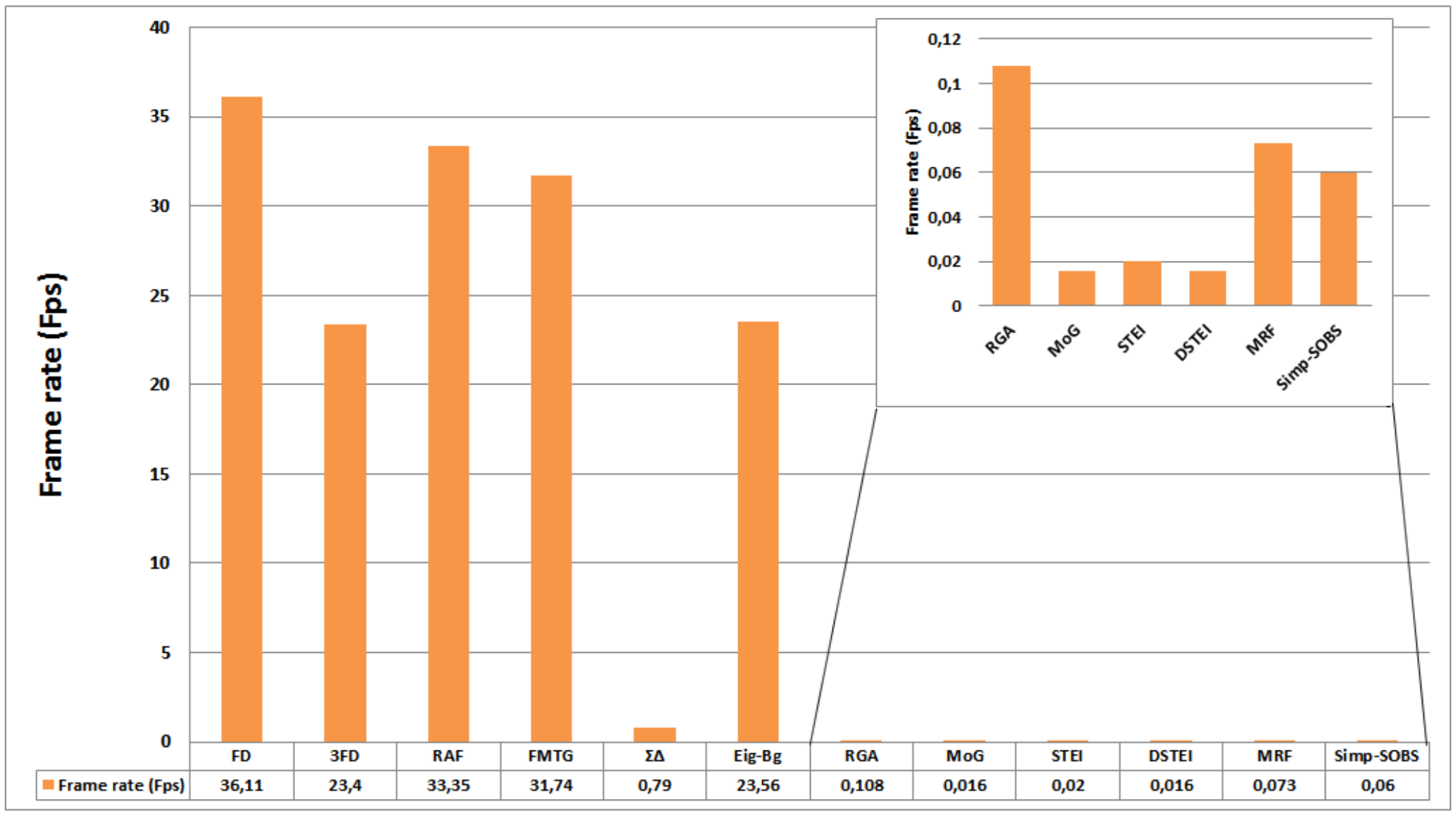

Fig. 18 Computational time for each method (presented in frames per second).

## 6 Conclusion

In this paper, we review and compare motion detection methods using one of the most recent, complete, and challenging datasets: CDnet2012 and CDnet2014. Detailed pixel evaluation was performed using different metrics to enable a user to determine the appropriate method for his or her needs.

From the results reported, we can conclude that there is no ideal method for all situations; each method performs well in some cases and fails in others. However, it is worth mentioning here that methods based on frame differences are not really designed to detect a complete silhouette of the moving object and thus are underrated here; they aim to detect motion and typically must be combined with other methods to achieve full segmentation. If we had to choose two methods based on the different challenging categories, they would be the simplified self-organized background subtraction[24] (Simp-SOBS) and the running Gaussian average[14] (RGA) methods; the former for the 'Bad weather', 'Baseline', 'intermittent object motion', 'low framerate', 'night video', 'shadow' and 'thermal' categories, and the latter for the 'PTZ', 'camera jitter', 'dynamic background' and 'turbulence' categories. This choice is justified by the high ranks of the methods for nearly all categories based on the average rank across categories as well as their acceptable execution times. The good performance of Simp-SOBS can be explained by the simple but powerful competitive learning used in SOM with an appropriate HSV color space that separates chromaticity from brightness information. The surprising results of RGA are linked to its low complexity with only a few parameters to adjust (e.g. compared to MoG[15]) that was sufficient for most areas of the images tested (only small image portions would have required more complex methods such as MoG), because the whole image background is not always dynamic except for the bad weather condition or maritime applications, where we found that the MoG is superior to the RGA in the tested categories). In the future, we will test other methods in order to expand the scope of this study and provide users with a complete benchmark of motion detection methods.

**Kamal Sehairi** is an Assistant Lecturer at École Normale Supérieure, Laghouat (ENS-L). He received his BS degree and Engineering degree in Electronics from the Amar Telidji Laghouat University in 2004 and 2009, respectively, and his Magister degree in Advanced Techniques in Signal Processing from École Militaire Polytechnique, Algiers, in 2012. He is a PhD student at Amar Telidji Laghouat University and a member of the LTSS laboratory. His current research interests include image processing and video processing, real-time implementation, FPGAs, classification and recognition, and video surveillance systems. He is a member of SPIE & IEEE.

**Fatima Chouireb** received her Electrical Engineering diploma in 1992, and her Magister and Ph.D. degrees from Saad Dahlab University of Blida, Algeria in 1996 and 2007 respectively.
In 1997, she joined the Electrical Engineering Department, Laghouat University, Algeria, as an Assistant Lecturer. Since Decembre 2007, Dr. Chouireb is an Assistant Professor at the same Department. She is also a Team Leader of ‘‘Signal, Image and Speech” research group of LTSS Laboratory, Laghouat University, Algeria. Her current research interests include: Signal, Image and Speech processing, computer vision, Localization/Mapping/SLAM and mobile robotics problems.

**Jean Meunier** received the B.S. degree in physics from the Université de Montréal, Montréal, QC, Canada, in 1981, the M.Sc.A. degree in applied mathematics in 1983, and the Ph.D. degree in biomedical engineering from the Ecole Polytechnique de Montréal, Montréal, in 1989. In 1989, after post-doctoral studies with the Montreal Heart Institute, Montréal, he joined the Department of Computer Science and Operations Research, Université de Montréal, where he is currently a Full Professor. He is a regular member of the Biomedical Engineering Institute at the same institution. His current research interests include computer vision and its applications to medical imaging and health care.

## Appendix

**Table 1** Pixel-based evaluation of different motion detection methods applied to the "PTZ" category

| | Recall | Specificity | FPR | FNR | PWC | Precision | F-Measure | $RM_c$ |
|---|---|---|---|---|---|---|---|---|
| GMM | 0,44955 | 0,98300 | 0,01700 | 0,55045 | 2,18306 | 0,26124 | **0,28955** | **3,57143** |
| RGA | 0,29935 | 0,98946 | 0,01054 | 0,70065 | **1,71521** | **0,29488** | 0,24364 | 3,71429 |
| RAF | 0,55135 | 0,96246 | 0,03754 | 0,44865 | 4,10254 | 0,18547 | 0,24236 | 4,42857 |
| 3FD | 0,14254 | **0,99055** | **0,00945** | 0,85746 | 1,74923 | 0,21983 | 0,12676 | 5,00000 |
| FD | 0,42000 | 0,97292 | 0,02708 | 0,58000 | 3,23362 | 0,21225 | 0,22099 | 5,28571 |
| FMTG | **0,67296** | 0,94991 | 0,05009 | **0,32704** | 5,22171 | 0,16603 | 0,24073 | 5,85714 |
| DSTEI | 0,45699 | 0,96035 | 0,03965 | 0,54301 | 4,42049 | 0,18210 | 0,22979 | 5,85714 |
| MRFMD | 0,13974 | 0,98662 | 0,01338 | 0,86026 | 2,14460 | 0,19404 | 0,12624 | 6,42857 |
| Simp-SOBS | 0,46153 | 0,93149 | 0,06851 | 0,53847 | 7,29140 | 0,13466 | 0,16815 | 7,42857 |
| Eig-Bg | 0,35897 | 0,80386 | 0,19614 | 0,64103 | 19,98432 | 0,03295 | 0,05791 | 9,71429 |
| ΣΔ | 0,07621 | 0,95886 | 0,04114 | 0,92379 | 4,92846 | 0,03709 | 0,04640 | 9,85714 |
| STEI | 0,32390 | 0,78885 | 0,21115 | 0,67610 | 21,53317 | 0,01634 | 0,03047 | 10,85714 |

**Table 2** Pixel-based evaluation of different motion detection methods applied to the "bad weather" category

| | Recall | Specificity | FPR | FNR | PWC | Precision | F-Measure | $RM_c$ |
|---|---|---|---|---|---|---|---|---|
| Simp-SOBS | 0,58117 | 0,99862 | 0,00138 | **0,41883** | **0,87063** | **0,88652** | **0,69965** | **2,14286** |
| GMM | 0,12651 | **0,99966** | **0,00034** | 0,87349 | 1,67592 | 0,81834 | 0,21127 | 4,57143 |
| FMTG | 0,25595 | 0,99546 | 0,00454 | 0,74405 | 1,89175 | 0,70015 | 0,35741 | 5,00000 |
| RAF | 0,24383 | 0,99563 | 0,00437 | 0,75617 | 1,94594 | 0,65861 | 0,35216 | 5,57143 |
| Eig-Bg | **0,60037** | 0,97762 | 0,02238 | 0,39963 | 2,91987 | 0,49428 | 0,50623 | 6,57143 |
| MRFMD | 0,04452 | 0,99863 | 0,00137 | 0,95548 | 2,07388 | 0,67153 | 0,08175 | 6,85714 |
| RGA | 0,32721 | 0,99209 | 0,00791 | 0,67279 | 2,19833 | 0,42407 | 0,33892 | 7,14286 |
| FD | 0,12061 | 0,99620 | 0,00380 | 0,87939 | 2,18648 | 0,62279 | 0,19972 | 7,57143 |
| DSTEI | 0,15961 | 0,99509 | 0,00491 | 0,84039 | 2,11153 | 0,61012 | 0,24112 | 7,57143 |
| 3 FD | 0,03300 | 0,99895 | 0,00105 | 0,96700 | 2,09177 | 0,60550 | 0,06235 | 7,71429 |
| ΣΔ | 0,27268 | 0,98994 | 0,01006 | 0,72732 | 2,61408 | 0,33206 | 0,25832 | 8,14286 |
| STEI | 0,39261 | 0,84577 | 0,15423 | 0,60739 | 15,89976 | 0,10713 | 0,14777 | 9,14286 |

**Table 3** Pixel-based evaluation of different motion detection methods applied to the "baseline" category

| | Recall | Specificity | FPR | FNR | PWC | Precision | F-Measure | $RM_c$ |
|---|---|---|---|---|---|---|---|---|
| Eig-Bg | **0,70521** | 0,99832 | 0,00168 | **0,29479** | **1,32715** | **0,93552** | **0,80366** | **2,14286** |
| Simp-SOBS | 0,48942 | 0,99649 | 0,00351 | 0,51058 | 2,63660 | 0,84190 | 0,60852 | 3,85714 |
| ΣΔ | 0,17359 | 0,99946 | 0,00054 | 0,82641 | 3,23513 | 0,90046 | 0,27642 | 5,42857 |
| RGA | 0,37160 | 0,99672 | 0,00328 | 0,62840 | 3,39816 | 0,69864 | 0,44997 | 5,85714 |
| FMTG | 0,41467 | 0,99483 | 0,00517 | 0,58533 | 3,00519 | 0,67255 | 0,45341 | 6,57143 |
| FD | 0,16392 | 0,99883 | 0,00117 | 0,83608 | 3,35138 | 0,79205 | 0,25216 | 6,71429 |
| RAF | 0,33089 | 0,99618 | 0,00382 | 0,66911 | 3,21227 | 0,68705 | 0,40236 | 7,14286 |
| DSTEI | 0,30211 | 0,99629 | 0,00371 | 0,69789 | 3,09809 | 0,68536 | 0,36369 | 7,42857 |
| 3FD | 0,06106 | **0,99947** | **0,00053** | 0,93894 | 3,61961 | 0,77192 | 0,10796 | 7,71429 |
| GMM | 0,29563 | 0,99640 | 0,00360 | 0,70437 | 3,30918 | 0,66375 | 0,39360 | 8,00000 |
| MRFMD | 0,11522 | 0,99872 | 0,00128 | 0,88478 | 3,51801 | 0,69588 | 0,18603 | 8,28571 |
| STEI | 0,60964 | 0,76918 | 0,23082 | 0,39036 | 23,14295 | 0,25211 | 0,20983 | 8,85714 |

**Table 4** Pixel-based evaluation of different motion detection methods applied to the "Camera Jitter" category

| | Recall | Specificity | FPR | FNR | PWC | Precision | F-Measure | $RM_c$ |
|---|---|---|---|---|---|---|---|---|
| Eig-Bg | **0,6076** | 0,9421 | 0,0579 | **0,3924** | 7,1837 | 0,3663 | **0,4391** | **3,1429** |
| RGA | 0,2905 | 0,9894 | 0,0106 | 0,7095 | 3,8901 | 0,5399 | 0,3741 | 3,5714 |
| GMM | 0,2374 | **0,9944** | **0,0056** | 0,7626 | **3,6016** | **0,6779** | 0,3344 | 3,7143 |
| Simp-SOBS | 0,5808 | 0,9373 | 0,0627 | 0,4192 | 7,7127 | 0,3411 | 0,4147 | 4,1429 |
| RAF | 0,4378 | 0,8926 | 0,1074 | 0,5622 | 12,6178 | 0,1664 | 0,2351 | 6,8571 |
| FMTG | 0,5246 | 0,8533 | 0,1467 | 0,4754 | 16,0025 | 0,1457 | 0,2236 | 7,0000 |
| ΣΔ | 0,0637 | 0,9837 | 0,0163 | 0,9363 | 5,5428 | 0,1953 | 0,0898 | 7,0000 |
| FD | 0,2809 | 0,9061 | 0,0939 | 0,7191 | 12,0018 | 0,1227 | 0,1668 | 8,1429 |
| 3FD | 0,1188 | 0,9513 | 0,0487 | 0,8812 | 8,4001 | 0,1040 | 0,1065 | 8,2857 |
| STEI | 0,4609 | 0,8205 | 0,1795 | 0,5391 | 19,6615 | 0,1436 | 0,1872 | 8,4286 |
| MRFMD | 0,0987 | 0,9554 | 0,0446 | 0,9013 | 8,1616 | 0,0888 | 0,0824 | 8,5714 |
| DSTEI | 0,4354 | 0,8314 | 0,1686 | 0,5646 | 18,5179 | 0,1041 | 0,1648 | 9,1429 |

**Table 5** Pixel-based evaluation of different motion detection methods applied to the “Dynamic Background” category

| | Recall | Specificity | FPR | FNR | PWC | Precision | F-Measure | $RM_c$ |
|---|---|---|---|---|---|---|---|---|
| RGA | 0,32604 | 0,99739 | **0,00261** | 0,67396 | **1,25771** | **0,44585** | **0,30555** | **3,00000** |
| DSTEI | 0,34040 | 0,96264 | 0,03736 | 0,65960 | 4,53876 | 0,21356 | 0,18193 | 5,00000 |
| GMM | 0,07816 | **0,99884** | 0,00116 | 0,92184 | 1,28203 | 0,43209 | 0,11900 | 5,28571 |
| Simp-SOBS | 0,52337 | 0,92515 | 0,07485 | 0,47663 | 7,89821 | 0,10130 | 0,16129 | 5,57143 |
| MRFMD | 0,11941 | 0,98458 | 0,01542 | 0,88059 | 2,65240 | 0,15423 | 0,09444 | 6,14286 |
| FMTG | 0,46085 | 0,90959 | 0,09041 | 0,53915 | 9,53249 | 0,12996 | 0,16795 | 6,42857 |
| RAF | 0,35967 | 0,91939 | 0,08061 | 0,64033 | 8,73891 | 0,09714 | 0,13346 | 7,14286 |
| 3FD | 0,09905 | 0,98135 | 0,01865 | 0,90095 | 2,95550 | 0,17437 | 0,08254 | 7,28571 |
| Eig-Bg | **0,62761** | 0,89217 | 0,10783 | **0,37239** | 11,02858 | 0,07684 | 0,13219 | 7,28571 |
| ΣΔ | 0,13895 | 0,98350 | 0,01650 | 0,86105 | 2,75199 | 0,07037 | 0,07060 | 7,57143 |
| FD | 0,22397 | 0,95006 | 0,04994 | 0,77603 | 5,86099 | 0,09671 | 0,10502 | 7,71429 |
| STEI | 0,42841 | 0,71230 | 0,28770 | 0,57159 | 29,06152 | 0,05091 | 0,07804 | 9,57143 |

**Table 6** Pixel-based evaluation of different motion detection methods applied to the “Intermittent object” category

| | Recall | Specificity | FPR | FNR | PWC | Precision | FMeasure | $RM_c$ |
|---|---|---|---|---|---|---|---|---|
| ΣΔ | 0,22742 | 0,99530 | 0,00470 | 0,77258 | 6,00318 | 0,68200 | **0,31732** | **4,28571** |
| Simp-SOBS | 0,21610 | 0,99506 | 0,00494 | 0,78390 | **5,81846** | 0,65376 | 0,30207 | 5,42857 |
| DSTEI | 0,09327 | 0,99691 | 0,00309 | 0,90673 | 6,23954 | 0,69406 | 0,15943 | 5,85714 |
| RGA | 0,12612 | 0,99609 | 0,00391 | 0,87388 | 6,33705 | 0,53360 | 0,18253 | 6,28571 |
| FD | 0,04566 | 0,99873 | 0,00127 | 0,95434 | 6,43023 | **0,73456** | 0,08313 | 6,42857 |
| FMTG | 0,11714 | 0,99452 | 0,00548 | 0,88286 | 6,18409 | 0,66480 | 0,19018 | 6,42857 |
| RAF | 0,10976 | 0,99541 | 0,00459 | 0,89024 | 6,31262 | 0,65421 | 0,18066 | 6,71429 |
| Eig-Bg | **0,63616** | 0,87557 | 0,12443 | **0,36384** | 14,03169 | 0,46714 | 0,45021 | 7,00000 |
| 3FD | 0,01516 | **0,99954** | 0,00046 | 0,98484 | 6,62883 | 0,72605 | 0,02929 | 7,14286 |
| MRFMD | 0,03628 | 0,99878 | **0,00122** | 0,96372 | 6,57922 | 0,67419 | 0,06807 | 7,28571 |
| GMM | 0,06211 | 0,99820 | 0,00180 | 0,93789 | 6,42132 | 0,58364 | 0,10818 | 7,28571 |
| STEI | 0,32070 | 0,90380 | 0,09620 | 0,67930 | 14,15520 | 0,27958 | 0,18753 | 7,85714 |

**Table 7** Pixel-based evaluation of different motion detection methods applied to the "Low Framerate" category

| | Recall | Specificity | FPR | FNR | PWC | Precision | F-Measure | $RM_c$ |
|---|---|---|---|---|---|---|---|---|
| Simp-SOBS | 0,51954 | 0,98959 | 0,01041 | 0,48046 | 2,56486 | **0,66767** | 0,44959 | **2,71429** |
| RGA | 0,40264 | 0,99610 | 0,00390 | 0,59736 | 2,58883 | 0,59523 | 0,36304 | 3,71429 |
| GMM | 0,33327 | **0,99736** | **0,00264** | 0,66673 | **2,31369** | 0,65832 | 0,35273 | 3,71429 |
| RAF | 0,51335 | 0,98607 | 0,01393 | 0,48665 | 2,90861 | 0,51617 | 0,40144 | 4,57143 |
| Eig-Bg | 0,56639 | 0,94749 | 0,05251 | 0,43361 | 6,54513 | 0,54566 | **0,46997** | 5,85714 |
| FD | 0,45044 | 0,97505 | 0,02495 | 0,54956 | 4,09714 | 0,34237 | 0,31730 | 6,42857 |
| FMTG | **0,57796** | 0,95039 | 0,04961 | **0,42204** | 6,06386 | 0,23980 | 0,28403 | 6,57143 |
| ΣΔ | 0,08207 | 0,99134 | 0,00866 | 0,91793 | 3,90652 | 0,50163 | 0,10016 | 7,14286 |
| MRFMD | 0,08464 | 0,98657 | 0,01343 | 0,91536 | 4,31021 | 0,11816 | 0,06137 | 8,85714 |
| 3FD | 0,10544 | 0,98281 | 0,01719 | 0,89456 | 4,58337 | 0,11600 | 0,08011 | 9,14286 |
| DSTEI | 0,31088 | 0,95063 | 0,04937 | 0,68912 | 6,88456 | 0,13012 | 0,15748 | 9,28571 |
| STEI | 0,34130 | 0,79972 | 0,20028 | 0,65870 | 22,17830 | 0,18730 | 0,06510 | 10,00000 |

**Table 8** Pixel-based evaluation of different motion detection methods applied to the "Night video" category

| | Recall | Specificity | FPR | FNR | PWC | Precision | F-Measure | $RM_c$ |
|---|---|---|---|---|---|---|---|---|
| FD | 0,26524 | 0,99547 | 0,00453 | 0,73476 | **2,01185** | 0,56769 | 0,35088 | **4,71429** |
| DSTEI | 0,40097 | 0,99043 | 0,00957 | 0,59903 | 2,24922 | 0,48116 | 0,42043 | 5,14286 |
| 3 FD | 0,06949 | **0,99921** | **0,00079** | 0,93051 | 2,03676 | **0,70609** | 0,12229 | 5,28571 |
| FMTG | **0,56808** | 0,98238 | 0,01762 | **0,43192** | 2,63505 | 0,45533 | **0,46018** | 5,28571 |
| Simp-SOBS | 0,47254 | 0,98834 | 0,01166 | 0,52746 | 2,29129 | 0,44207 | 0,43995 | 5,28571 |
| MRFMD | 0,07002 | 0,99876 | 0,00124 | 0,92998 | 2,07952 | 0,56291 | 0,12237 | 5,57143 |
| RAF | 0,50793 | 0,98401 | 0,01599 | 0,49207 | 2,63158 | 0,42673 | 0,43843 | 5,71429 |
| GMM | 0,20595 | 0,99388 | 0,00612 | 0,79405 | 2,26189 | 0,39635 | 0,26142 | 7,00000 |
| Eig-Bg | 0,48550 | 0,97250 | 0,02750 | 0,51450 | 3,77828 | 0,31558 | 0,35541 | 7,71429 |
| ΣΔ | 0,03625 | 0,99749 | 0,00251 | 0,96375 | 2,27967 | 0,38100 | 0,05685 | 8,14286 |
| RGA | 0,32350 | 0,98592 | 0,01408 | 0,67650 | 2,84418 | 0,30709 | 0,29126 | 8,28571 |
| STEI | 0,45413 | 0,77635 | 0,22365 | 0,54587 | 22,85890 | 0,05566 | 0,09322 | 9,85714 |

**Table 9** Pixel-based evaluation of different motion detection methods applied to the "Shadow" category

| | Recall | Specificity | FPR | FNR | PWC | Precision | F-Measure | $RM_c$ |
|---|---|---|---|---|---|---|---|---|
| Simp-SOBS | 0,50198 | 0,99044 | 0,00956 | 0,49802 | **3,07260** | 0,75207 | **0,59353** | **4,28571** |
| RGA | 0,32784 | 0,99372 | 0,00628 | 0,67216 | 3,61823 | 0,71265 | 0,43611 | 4,57143 |
| GMM | 0,24914 | 0,99518 | 0,00482 | 0,75086 | 3,61844 | 0,74109 | 0,34635 | 5,00000 |
| FD | 0,11852 | 0,99764 | 0,00236 | 0,88148 | 4,08282 | 0,71392 | 0,19762 | 6,14286 |
| DSTEI | 0,20761 | 0,99541 | 0,00459 | 0,79239 | 3,92670 | 0,67913 | 0,30582 | 6,28571 |
| FMTG | 0,30569 | 0,99083 | 0,00917 | 0,69431 | 3,94654 | 0,63377 | 0,39529 | 6,42857 |
| 3 FD | 0,03776 | **0,99962** | **0,00038** | 0,96224 | 4,24996 | **0,77620** | 0,07113 | 6,85714 |
| Eig-Bg | 0,61292 | 0,95488 | 0,04512 | 0,38708 | 6,08321 | 0,50480 | 0,50255 | 7,14286 |
| MRFMD | 0,06100 | 0,99894 | 0,00106 | 0,93900 | 4,24903 | 0,69451 | 0,11048 | 7,28571 |
| RAF | 0,24641 | 0,99171 | 0,00829 | 0,75359 | 4,09855 | 0,62128 | 0,33949 | 7,42857 |
| ΣΔ | 0,07324 | 0,99648 | 0,00352 | 0,92676 | 4,39514 | 0,51865 | 0,12251 | 8,28571 |
| STEI | **0,71537** | 0,76417 | 0,23583 | **0,28463** | 23,76070 | 0,12635 | 0,20898 | 8,28571 |

**Table 10** Pixel-based evaluation of different motion detection methods applied to the "Thermal" category

| | Recall | Specificity | FPR | FNR | PWC | Precision | F-Measure | $RM_c$ |
|---|---|---|---|---|---|---|---|---|
| Eig-Bg | **0,57527** | 0,98983 | 0,01017 | **0,42473** | **3,34477** | 0,75660 | **0,63940** | **4,71429** |
| RAF | 0,16894 | 0,99785 | 0,00215 | 0,83106 | 6,46159 | 0,77678 | 0,24840 | 4,85714 |
| FD | 0,08293 | 0,99952 | 0,00048 | 0,91707 | 6,76412 | **0,89736** | 0,13254 | 5,28571 |
| FMTG | 0,17280 | 0,99771 | 0,00229 | 0,82720 | 6,53223 | 0,76712 | 0,23493 | 5,42857 |
| Simp-SOBS | 0,30052 | 0,99103 | 0,00897 | 0,69948 | 6,14208 | 0,72833 | 0,40935 | 5,42857 |
| DSTEI | 0,09116 | 0,99884 | 0,00116 | 0,90884 | 6,80556 | 0,77026 | 0,13743 | 6,00000 |
| 3FD | 0,03946 | 0,99975 | 0,00025 | 0,96054 | 6,90147 | 0,87329 | 0,06998 | 6,42857 |
| RGA | 0,14817 | 0,99652 | 0,00348 | 0,85183 | 6,69029 | 0,66375 | 0,23215 | 7,14286 |
| MRFMD | **0,01977** | **0,99979** | **0,00021** | 0,98023 | 6,97932 | 0,79833 | 0,03642 | 7,28571 |
| GMM | 0,07448 | 0,99857 | 0,00143 | 0,92552 | 6,79791 | 0,72397 | 0,13242 | 7,57143 |
| STEI | 0,41382 | 0,65387 | 0,34613 | 0,58618 | 33,18293 | 0,09735 | 0,15688 | 8,28571 |
| ΣΔ | 0,02021 | 0,99792 | 0,00208 | 0,97979 | 7,12201 | 0,40127 | 0,03848 | 9,57143 |

**Table 11** Pixel-based evaluation of different motion detection methods applied to the “Turbulence” category

| | Recall | Specificity | FPR | FNR | PWC | Precision | F-Measure | $RM_c$ |
|---|---|---|---|---|---|---|---|---|
| RGA | 0,37053 | 0,99522 | 0,00478 | 0,62947 | 0,89475 | 0,22000 | 0,24396 | **3,42857** |
| GMM | 0,15444 | **0,99975** | **0,00025** | 0,84556 | **0,46990** | **0,75562** | **0,24731** | 4,14286 |
| Eig-Bg | **0,78752** | 0,95435 | 0,04565 | **0,21248** | 4,71277 | 0,10395 | 0,15637 | 4,71429 |
| Simp-SOBS | 0,78287 | 0,95072 | 0,04928 | 0,21713 | 5,07985 | 0,11308 | 0,16392 | 5,28571 |
| DSTEI | 0,46520 | 0,97167 | 0,02833 | 0,53480 | 3,17950 | 0,05695 | 0,09098 | 5,71429 |
| ΣΔ | 0,34950 | 0,97966 | 0,02034 | 0,65050 | 2,48953 | 0,07998 | 0,08646 | 5,85714 |
| RAF | 0,50183 | 0,95535 | 0,04465 | 0,49817 | 4,79166 | 0,06753 | 0,09773 | 6,14286 |
| MRFMD | 0,16698 | 0,98938 | 0,01062 | 0,83302 | 1,56510 | 0,04329 | 0,05261 | 7,00000 |
| FMTG | 0,59875 | 0,91203 | 0,08797 | 0,40125 | 9,05480 | 0,05586 | 0,08904 | 7,28571 |
| 3FD | 0,17101 | 0,96715 | 0,03285 | 0,82899 | 3,76600 | 0,03516 | 0,04327 | 8,57143 |
| STEI | 0,58499 | 0,81651 | 0,18349 | 0,41501 | 18,58043 | 0,03174 | 0,05193 | 9,28571 |
| FD | 0,33346 | 0,90664 | 0,09336 | 0,66654 | 9,71757 | 0,02896 | 0,04452 | 10,57143 |